%% file: Main_uav_ei.tex
\documentclass[lettersize,journal]{IEEEtran}
\usepackage{amsmath,amsfonts}
\usepackage{algorithmic}
\usepackage{algorithm}
\usepackage{array}
\usepackage[caption=false,font=normalsize,labelfont=sf,textfont=sf]{subfig}
\usepackage{textcomp}
\usepackage{url}
\usepackage{verbatim}
\usepackage{graphicx}
\usepackage{cite}
\usepackage{booktabs}
\usepackage{color}
\usepackage{multirow}
\usepackage{hyperref}
\makeatletter

\def\section{\@startsection{section}{1}{\z@}{1.5ex plus 1.5ex minus 0.5ex}%
{0.7ex plus 1ex minus 0.5ex}{\normalfont\Large\bfseries}}

\def\subsection{\@startsection{subsection}{2}{\z@}{1.5ex plus 1.5ex minus 0.5ex}%
{0.7ex plus .5ex minus 0.5ex}{\normalfont\large\bfseries}}

\def\subsubsection{\@startsection{subsubsection}{3}{\z@}{1.5ex plus 1.5ex minus 0.5ex}%
{0.5ex plus .5ex minus 0.5ex}{\normalfont\normalsize\bfseries}}

\def\@seccntformat#1{\csname the#1\endcsname\quad}

\def\l@section#1#2{\addpenalty{\@secpenalty}\addvspace{1.0em plus 1pt}%
\@tempdima 1.5em \begingroup
\parindent \z@ \rightskip \@pnumwidth \parfillskip -\@pnumwidth
\bfseries \leavevmode #1\hfil \hbox to\@pnumwidth{\hss #2}\par
\endgroup}
\makeatother

\usepackage{booktabs}
\usepackage{tabularx}
\usepackage{array}

\newcolumntype{Y}{>{\raggedright\arraybackslash}X}

\begin{document}

\title{UAVs Meet Embodied Intelligence: \\Bridging Human Intents and Flying Dynamics \\Via Harnessing Physical-Digital AI Agents}
    
\author{Yonglin Tian\textsuperscript{1}, Weiyi Wang\textsuperscript{1},
Houhua Lu\textsuperscript{2}, Xinyi Li\textsuperscript{3},
Yihao Wu\textsuperscript{1}, Jingyang Chen\textsuperscript{4},
Jianli Sun\textsuperscript{1}, Chengxiang Li\textsuperscript{5},
Yinuo Chen\textsuperscript{6}, Fei Lin\textsuperscript{7},
Tengchao Zhang\textsuperscript{7}, Jing Yang\textsuperscript{8},
Deyi Ji\textsuperscript{9}, Jian Di\textsuperscript{10},
Naiqi Wu\textsuperscript{7}, and Yisheng Lv\textsuperscript{1}
\thanks{\begin{tabular}{@{}r@{\quad}>{\raggedright\arraybackslash}p{0.84\columnwidth}@{}}
1 & Institute of Automation, Chinese Academy of Sciences, China.\\
2 & Beihang University, China.\\
3 & Shanxi Datong University, China.\\
4 & Johns Hopkins University, USA.\\
5 & University of Sanya, China.\\
6 & Wenzhou-Kean University, China.\\
7 & Macau University of Science and Technology, Macao, China.\\
8 & People's Public Security University of China, China.\\
9 & KOKONI 3D, Moxin Technology, China.\\
10 & University of Science and Technology of China, China.
\end{tabular}\\
}
}

\markboth{}%
{Shell \MakeLowercase{\textit{et al.}}: UAV Embodied Intelligence: Bridging Human Intent and World Dynamics}


\maketitle

\begin{abstract}
Unmanned aerial vehicles (UAVs) extend embodied intelligence into continuous three-dimensional space, where perception, reasoning, physical embodiment, and action are tightly coupled through flight and environmental interaction. Recent advances in foundation models, world models, and AI agents are shifting UAV autonomy from task-specific perception and control toward systems that can interpret human intent, understand open environments, reason about physical consequences, and organize complex behaviors under embodiment and flight-dynamic constraints. We characterize this emerging paradigm as \emph{UAV embodied intelligence (UAV EI)} and distinguish it from its system realization, the \emph{embodied-intelligent UAV (EI UAV)}. To provide a unified view of the field, we introduce a 5+5 framework that describes UAV EI through five capability dimensions and EI UAVs through five architectural layers spanning physical embodiment, general cognition, embodied skills, external interaction, and system harnessing. Based on this framework, we systematically review recent progress in embodied morphology, embodied perception, world models, embodied planning, vision-language navigation, embodied manipulation, and embodied collaboration. We further identify long-horizon autonomy, predictive physical reasoning, test-time skill acquisition, and autonomous capability evolution as key challenges toward more general aerial embodied intelligence. Finally, we argue that harnessing \emph{physical-digital AI agents}, through persistent coupling of digital intelligence with physical sensing, dynamics, action, and feedback, provides a system-level pathway toward adaptive and continuously evolving UAV autonomy. Project resources are available at our
\href{https://hub-tian.github.io/UAVs_Meet_Embodied-Intelligence/}{project website}
and
\href{https://github.com/Hub-Tian/UAVs_Meet_Embodied-Intelligence}{GitHub repository}.
\end{abstract}


\input{sections/introduction}

\input{sections/uav-ei-definition}

\input{sections/uav-embodied-morphology}

\input{sections/uav-embodied-perception}

\input{sections/uav-world-models}
\input{sections/uav-embodied-planning}
\input{sections/uav-vision-language-navigation}

\input{sections/uav-embodied-manipulation}
\input{sections/uav-embodied-cooperation}

\input{sections/conclusion}

\bibliographystyle{IEEEtran}
\bibliography{uav_ei_refs}

\vfill

\end{document}

%% file: sections/introduction.tex
\section{Introduction}

Flight extends robotic intelligence from predominantly planar mobility to active perception and interaction in continuous three-dimensional space. By changing position, altitude, attitude, and viewpoint, unmanned aerial vehicles (UAVs) can access otherwise unreachable regions, overcome occlusions, rapidly approach distant targets, and actively shape how observations are acquired. Their motion therefore influences not only where they can operate, but also what they can perceive and how they can interact with the physical world, making UAVs a distinctive platform for embodied intelligence.

UAV autonomy has progressed from flight control toward task-specific intelligence. Early research focused primarily on stabilization, trajectory tracking, and disturbance rejection, while advances in computer vision, deep learning, reinforcement learning, and autonomous navigation subsequently enabled increasingly autonomous perception, navigation, planning, and coordination \cite{bouguettaya2021vehicle,wu2021deep}. Despite this progress, most existing systems remain optimized for predefined objectives, closed-set observations, and fixed task interfaces. The UAV body, environment, and mission requirements are often treated as predetermined conditions rather than coupled components of intelligent behavior, limiting the ability to interpret open-ended human requirements, actively acquire task-relevant information, and reorganize behavior as missions evolve \cite{xiao2025uav,sun2026open}.

Recent advances in foundation models and artificial intelligence (AI) agents provide an opportunity to move beyond this task-specific paradigm \cite{tian2025uavs}. Natural language enables UAVs to interpret semantic objectives and implicit constraints beyond predefined commands, while general-purpose multimodal models support more transferable understanding of objects, scenes, spatial relations, and dynamic events. Advances in memory, reasoning, world modeling, and agent-based planning further enable long-horizon task decomposition, active information acquisition, prediction of action consequences, and adaptive decision-making. Their significance therefore lies not simply in improving individual modules, but in connecting human intent, embodied state, environment understanding, task reasoning, and physical action within a unified closed loop.

We refer to this emerging paradigm as UAV embodied intelligence (UAV EI). UAV EI emphasizes that intelligence must be continuously grounded in aerial embodiment, including what the UAV can perceive, how it can move or interact, and how the physical world responds to its actions. We characterize this coupling through physical-digital artificial intelligence (PD-AI) agents. The digital agent provides semantic understanding, memory, world modeling, reasoning, and task organization, whereas the physical agent realizes sensing, flight, control, actuation, and physical interaction. Physical observations and execution outcomes continuously ground digital reasoning, while digital decisions guide subsequent perception and action. Human instructions, preferences, and feedback further connect this physical-digital loop with social space, forming a cyber-physical-social system (CPSS) \cite{wang2010emergence}. UAV EI therefore concerns the persistent coupling of human intent, digital intelligence, physical embodiment, and the evolving environment.

To organize this rapidly developing field, we distinguish UAV EI, which describes the capabilities required for embodied aerial intelligence, from the embodied-intelligent UAV (EI UAV), which describes their system realization. Based on this distinction, we develop a 5+5 framework that connects five capability dimensions of UAV EI with five architectural layers of an EI UAV, providing a common basis for understanding research across embodiment, perception, world modeling, planning, navigation, manipulation, and collaboration. The main contributions of this survey are summarized as follows:

\begin{itemize}

\item We establish a unified 5+5 framework that distinguishes the capabilities of UAV EI from their realization through five architectural layers of an EI UAV.

\item We systematically organize UAV embodied intelligence across embodiment, perception, world models, planning, navigation, manipulation, and collaboration, emphasizing their coupling within the perception-reasoning-action loop.

\item We present PD-AI agents as a system-level perspective for UAV EI and identify long-horizon aerial autonomy, flight-grounded physical reasoning, changing flight conditions adaptation, and experience-driven capability evolution as key challenges toward more general aerial autonomy.

\end{itemize}

The remainder of this survey is organized as follows. Section~II introduces the UAV EI and EI UAV concepts and presents the 5+5 framework. Sections~III to VIII review UAV embodiment, embodied perception, world models, embodied planning, embodied navigation and manipulation, and embodied collaboration. Section~IX discusses the major challenges and future directions, including the role of PD-AI agents and the Harness Layer, followed by the conclusion in Section~X.

%% file: sections/uav-ei-definition.tex
\section{UAV EI and EI UAV}
\label{sec:uav_ei_eiuav}

\begin{figure*}[t]
    \centering
    \includegraphics[width=\textwidth]{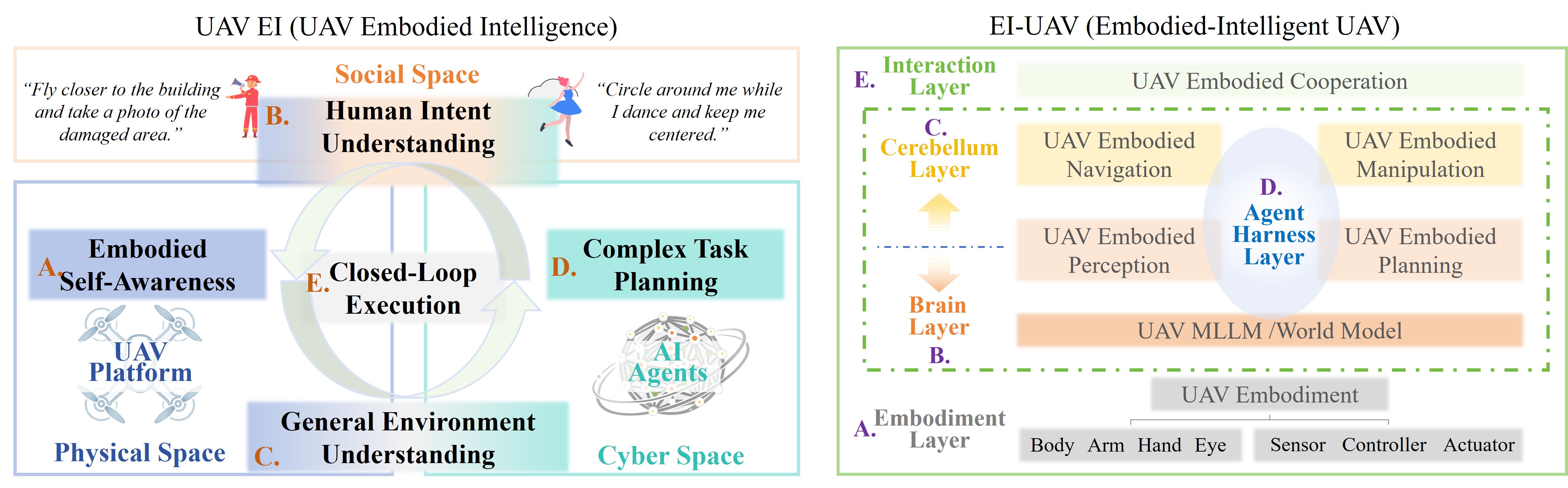}
    \caption{The 5+5 framework of UAV EI and EI UAV, comprising five capability dimensions and five architectural layers.}
    \label{fig:five_essentials}
\end{figure*}

UAV EI and EI UAV provide complementary capability and system perspectives on aerial embodied intelligence. UAV EI describes what an aerial embodied agent should be able to accomplish, whereas an EI UAV describes how these capabilities are realized in an operational system. Fig.~\ref{fig:five_essentials} summarizes this relationship through a 5+5 framework.

\subsection{UAV EI: Five Capability Dimensions}

UAV EI comprises five coupled capabilities: Human Intent Understanding, Embodied Self-Awareness, General Environment Understanding, Complex Task Planning, and Closed-Loop Execution.

Human Intent Understanding connects human objectives with autonomous behavior. An EI UAV should interpret natural-language instructions, semantic objectives, preferences, and constraints and transform them into actionable task representations, allowing users to specify what should be achieved without prescribing every execution step.

Embodied Self-Awareness grounds decisions in the UAV's own physical state and capabilities. Morphology, pose, motion state, sensing configuration, payload, energy, and actuation limits constrain what the platform can perceive and physically accomplish.

General Environment Understanding extends perception from task-specific recognition toward transferable semantic, spatial, geometric, and dynamic understanding. It is inherently active because the UAV can adjust position, altitude, viewpoint, trajectory, or sensing configuration to acquire task-relevant information.

Complex Task Planning transforms high-level objectives into physically feasible sequences of perception and action. It reasons jointly over task requirements, environmental states, embodiment constraints, action consequences, and available resources, while allowing intermediate objectives and strategies to change as the mission evolves.

Closed-Loop Execution grounds decisions in physical observations and execution outcomes. Feedback from the environment updates internal states and subsequent actions, coupling perception, reasoning, planning, and embodied execution beyond low-level feedback control.

These capabilities operate jointly rather than independently. UAV EI therefore depends on grounding intelligence in aerial embodiment and physical interaction rather than on any individual model or algorithm.

\subsection{EI UAV: Five Architectural Layers}

An EI UAV realizes these capabilities through five architectural layers: the Embodiment Layer, Brain Layer, Cerebellum Layer, Interaction Layer, and Harness Layer.

The Embodiment Layer provides the physical basis for perception and action, including the aerial body, sensors, controllers, actuators, computing devices, payloads, and manipulators when available. Its morphology, sensing geometry, flight dynamics, energy, and actuation mechanisms define the physical possibilities and constraints of the system.

The Brain Layer provides general semantic, predictive, and deliberative intelligence. Its core components include multimodal large language models (MLLMs) and world models, which support semantic understanding, multimodal grounding, task reasoning, state representation, and prediction of future consequences. These capabilities primarily support embodied perception and planning.

The Cerebellum Layer translates high-level decisions into embodiment-specific skills, primarily navigation and manipulation. These skills realize aerial mobility and physical interaction under geometric, dynamic, control, and temporal constraints.

The Interaction Layer enables cooperation with humans, other UAVs, ground robots, and heterogeneous embodied agents. By exchanging observations, intentions, task states, plans, and actions, it extends individual intelligence toward distributed embodied operation.

The Harness Layer provides the runtime infrastructure for coordinating models, memory, knowledge, tools, skills, context, and physical resources. Agent runtime and orchestration maintain task state and execution flow, while execution monitoring and guardrails support feedback, recovery, and constrained physical action. The Harness Layer therefore integrates otherwise separate capabilities into a coherent executable system.

These layers are tightly coupled rather than sequential. The Embodiment Layer defines physical constraints, the Brain Layer provides general cognition, the Cerebellum Layer realizes embodied skills, the Interaction Layer supports external cooperation, and the Harness Layer coordinates their operation. Together, they provide the system architecture for realizing UAV EI.

%% file: sections/uav-embodied-morphology.tex
\begin{figure*}[!t]
    \centering
    \includegraphics[width=\textwidth]{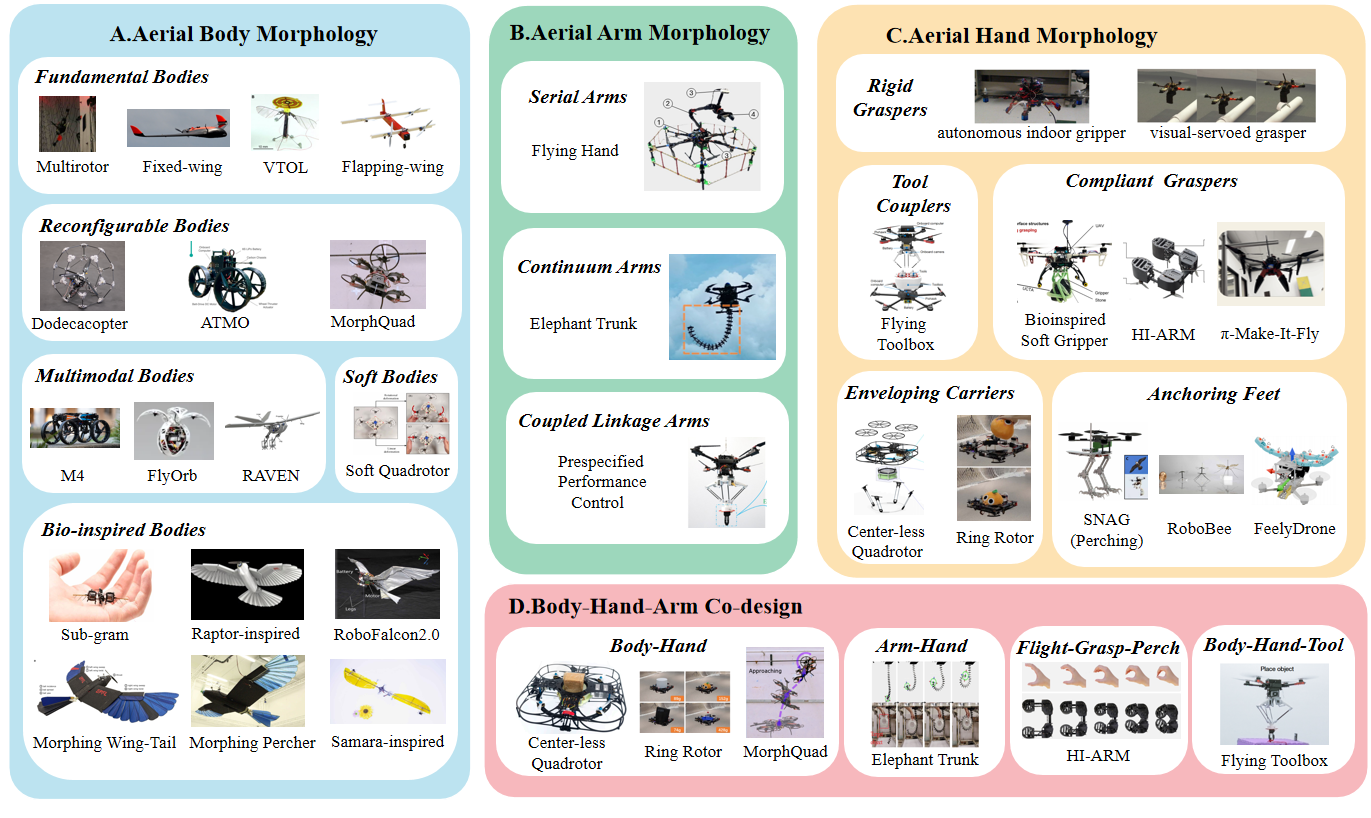}
    \caption{Overview of UAV embodied morphology across four categories.
    (A) Aerial body morphology includes fundamental bodies, including multirotor \cite{mellinger2011minimum}, fixed-wing \cite{bohn2019deep}, VTOL \cite{Muraoka2009}, and flapping-wing \cite{graule2016perching}; reconfigurable bodies, including Dodecacopter \cite{garanger2025dodecacopter}, ATMO \cite{mandralis2025atmo}, and MorphQuad \cite{pacheco2026morphquad}; multimodal bodies, including M4 \cite{sihite2023multi}, FlyOrb \cite{li2026flyorb}, and RAVEN \cite{shin2024fast}; soft bodies, represented by the soft quadrotor \cite{xu2026passive}; and bio-inspired bodies, including the sub-gram flapping-wing robot \cite{kim2025acrobatics}, raptor-inspired platform \cite{phan2024twist}, RoboFalcon 2.0 \cite{chen2025flapping}, morphing wing-tail platform \cite{jeger2024adaptive}, morphing percher \cite{wuest2024agile}, and Samara-inspired drone \cite{bai2022bioinspired}.
    (B) Aerial arm morphology includes serial arms, represented by Flying Hand \cite{he2025flying}; coupled linkage arms, represented by Prespecified Performance Control \cite{cao2025prespecified}; and continuum arms, represented by Aerial Elephant Trunk \cite{peng2025dexterous}.
    (C) Aerial hand morphology includes rigid graspers, including the autonomous indoor gripper \cite{ghadiok2011autonomous} and visual-servoed grasper \cite{thomas2014toward}; compliant graspers, including the bio-inspired soft gripper \cite{guo2024powerful}, HI-ARM \cite{wu2026hand}, and $\pi$-Make-It-Fly \cite{tucker2026pi}; enveloping carriers, including Center-less Quadrotor \cite{tan2025center} and Ring-Rotor \cite{wu2023ring}; anchoring feet, including SNAG \cite{roderick2021bird}, RoboBee \cite{hyun2025sticking}, and FeelyDrone \cite{bredenbeck2026aerial}; and tool couplers, represented by Flying Toolbox \cite{cao2025proximal}.
    (D) Body-arm-hand co-design includes body-hand integration in Center-less Quadrotor \cite{tan2025center}, Ring-Rotor \cite{wu2023ring}, and MorphQuad \cite{pacheco2026morphquad}; arm-hand integration in Aerial Elephant Trunk \cite{peng2025dexterous}; flight-grasp-perch integration in HI-ARM \cite{wu2026hand}; and body-hand-tool integration in Flying Toolbox \cite{cao2025proximal}.}
    \label{fig:aerial_morphology}
\end{figure*}

\section{UAV Embodied Morphology}

Embodiment defines how intelligence perceives, moves, and interacts with the physical world. For UAVs, morphology determines flight characteristics, sensing geometry, interaction capability, payload capacity, and physical affordances. As illustrated in Fig.~\ref{fig:aerial_morphology}, UAV morphology can be organized into three coupled levels: the body determines mobility and environmental adaptation, the arm extends interaction workspace, and the hand establishes physical contact with objects, surfaces, and tools.

\subsection{Aerial Body Morphology}

Aerial body morphology defines the physical basis of UAV mobility and interaction. Different structures introduce distinct trade-offs among hovering, endurance, maneuverability, accessibility, payload, and stability.

\textbf{Fundamental Aerial Bodies.}
Multirotors provide hovering capability and high maneuverability for inspection and interaction tasks \cite{mellinger2011minimum}. Fixed-wing platforms improve endurance and cruise efficiency but impose constraints on takeoff, landing, and low-speed flight \cite{bohn2019deep}. Hybrid vertical takeoff and landing (VTOL) platforms combine hovering and efficient forward flight through mechanisms such as tilt rotors \cite{Muraoka2009}, while flapping-wing platforms exploit lightweight bio-inspired mechanisms for agile flight \cite{graule2016perching}.

\textbf{Reconfigurable Bodies.}
Reconfigurable bodies modify geometry or actuation structures according to task or environmental requirements, while introducing changes in mass distribution, dynamics, and control. The Dodecacopter achieves modular three-dimensional actuation through a reconfigurable rotor structure \cite{garanger2025dodecacopter}. ATMO enables transitions between aerial and ground locomotion through morphological transformation \cite{mandralis2025atmo}. MorphQuad further introduces continuously articulated rotor modules for omnidirectional force generation and agile maneuvering \cite{pacheco2026morphquad}.

\textbf{Multimodal Locomotion Bodies.}
Multimodal bodies extend UAV mobility beyond flight by combining aerial motion with terrestrial locomotion. M4 integrates flight, rolling, crawling, and balancing within a unified platform \cite{sihite2023multi}. FlyOrb combines multirotor flight with passive rolling through a spherical structure \cite{li2026flyorb}, while RAVEN employs multifunctional legs for walking, hopping, obstacle traversal, and jump-assisted takeoff \cite{shin2024fast}.

\textbf{Bio-Inspired Bodies.}
Bio-inspired bodies incorporate functional principles from natural flyers to improve agility, efficiency, stability, or interaction. Sub-gram flapping-wing robots exploit lightweight structures for agile flight at small scales \cite{kim2025acrobatics}. Bird-inspired systems use tail and wing morphing mechanisms to improve maneuverability, takeoff, and perching \cite{phan2024twist,chen2025flapping,jeger2024adaptive,wuest2024agile}. Feather-based mechanoreception further demonstrates the integration of sensing into morphology \cite{li2024avian}, while Samara-inspired designs exploit passive autorotation for stable flight \cite{bai2022bioinspired}.

\textbf{Soft Bodies.}
Soft bodies introduce structural compliance to improve collision tolerance and adaptability in constrained environments. The Passive Morphing Soft Quadrotor deforms during collision or narrow-space traversal and recovers its original shape \cite{xu2026passive}. Such designs improve physical robustness but also modify structural dynamics and control requirements.

\subsection{Aerial Arm Morphology.}
Aerial arms extend UAV interaction beyond the vehicle body. Unlike fixed-base manipulators, aerial arms directly influence vehicle pose, center of mass, inertia, and flight stability. Existing designs include serial arms, coupled linkage arms, and continuum arms.

\textbf{Serial Robotic Arms.}
Serial robotic arms provide controllable end-effector position and orientation through articulated rigid links. The Flying Hand combines an aerial platform with an articulated arm for aerial manipulation tasks \cite{he2025flying}. Such designs provide high dexterity but introduce stronger flight-manipulation coupling.

\textbf{Coupled Linkage Arms.}
Coupled linkage arms use mechanically constrained structures instead of general-purpose serial chains. The Prespecified Performance Control framework represents this category by exploiting lightweight and predictable linkage mechanisms for aerial interaction \cite{cao2025prespecified}.

\textbf{Continuum Arms.}
Continuum arms provide adaptive interaction through continuous deformation, tendon actuation, or cable-driven bending. The Aerial Elephant Trunk employs a tendon-driven continuum structure for shape-adaptive reaching and wrapping interaction in cluttered environments \cite{peng2025dexterous}.

\subsection{Aerial Hand Morphology.}
Aerial hand morphology defines the physical interface between UAVs and objects, surfaces, payloads, or tools. Due to the limited payload and moving-base dynamics, aerial hands must balance grasp reliability, stability, and tolerance to uncertainty. Existing designs include rigid graspers, compliant graspers, enveloping carriers, anchoring feet, and tool couplers.

\textbf{Rigid Graspers.}
Rigid graspers provide predictable contact geometry and force transmission through stiff fingers, claws, hooks, or fixtures. Classical aerial grasping systems demonstrate this approach through autonomous and visual-servoed grasping mechanisms \cite{ghadiok2011autonomous,thomas2014toward}.

\textbf{Compliant Graspers.}
Compliant graspers use deformable or underactuated structures to tolerate object variation and pose uncertainty. Bio-inspired soft graspers enable adaptive grasping through passive deformation \cite{guo2024powerful}. HI-ARM introduces a tendon-driven aerial hand supporting multiple grasping modes \cite{wu2026hand}, while lightweight compliant end-effectors support dynamic aerial pick-and-place under payload disturbance \cite{tucker2026pi}.

\textbf{Enveloping Carriers.}
Enveloping carriers stabilize payloads through structural containment rather than localized grasping. The Center-less Quadrotor integrates a soft carrier near the vehicle center \cite{tan2025center}, while Ring-Rotor uses a retractable ring structure for object enclosure and transportation \cite{wu2023ring}.

\textbf{Anchoring Feet.}
Anchoring feet enable landing, perching, and persistent physical attachment. SNAG achieves bird-inspired perching through dynamic grasping and tendon locking mechanisms \cite{roderick2021bird}. RoboBee explores insect-inspired landing structures for micro aerial robots \cite{hyun2025sticking}, while tactile-enabled compliant hands improve aerial perching under pose uncertainty \cite{bredenbeck2026aerial}.

\textbf{Tool Couplers.}
Tool couplers provide interfaces for docking, exchanging, and operating task-specific tools. Flying Toolbox demonstrates cooperative aerial tool transport and exchange between specialized UAVs \cite{cao2025proximal}.

\subsection{Body-Arm-Hand Co-Design}

Body, arm, and hand morphologies are strongly coupled because payload placement, arm motion, and contact forces directly affect vehicle dynamics. Body-hand integration is demonstrated by Ring-Rotor and the Center-less Quadrotor, where payload containment is directly embedded into the airframe structure \cite{wu2023ring,tan2025center}. MorphQuad further uses body-level thrust vectoring to generate interaction forces without conventional manipulators \cite{pacheco2026morphquad}.

Arm-hand integration combines reaching and contact functions. The Aerial Elephant Trunk integrates continuum reaching and wrapping interaction within a unified structure \cite{peng2025dexterous}, while HI-ARM combines grasping and perching through a compliant aerial hand design \cite{wu2026hand}. Flying Toolbox extends co-design across multiple UAVs by coupling tool carrying and operation through dedicated interfaces \cite{cao2025proximal}.

%% file: sections/uav-embodied-perception.tex
\begin{figure*}[t]
\centering
\includegraphics[width=\textwidth]{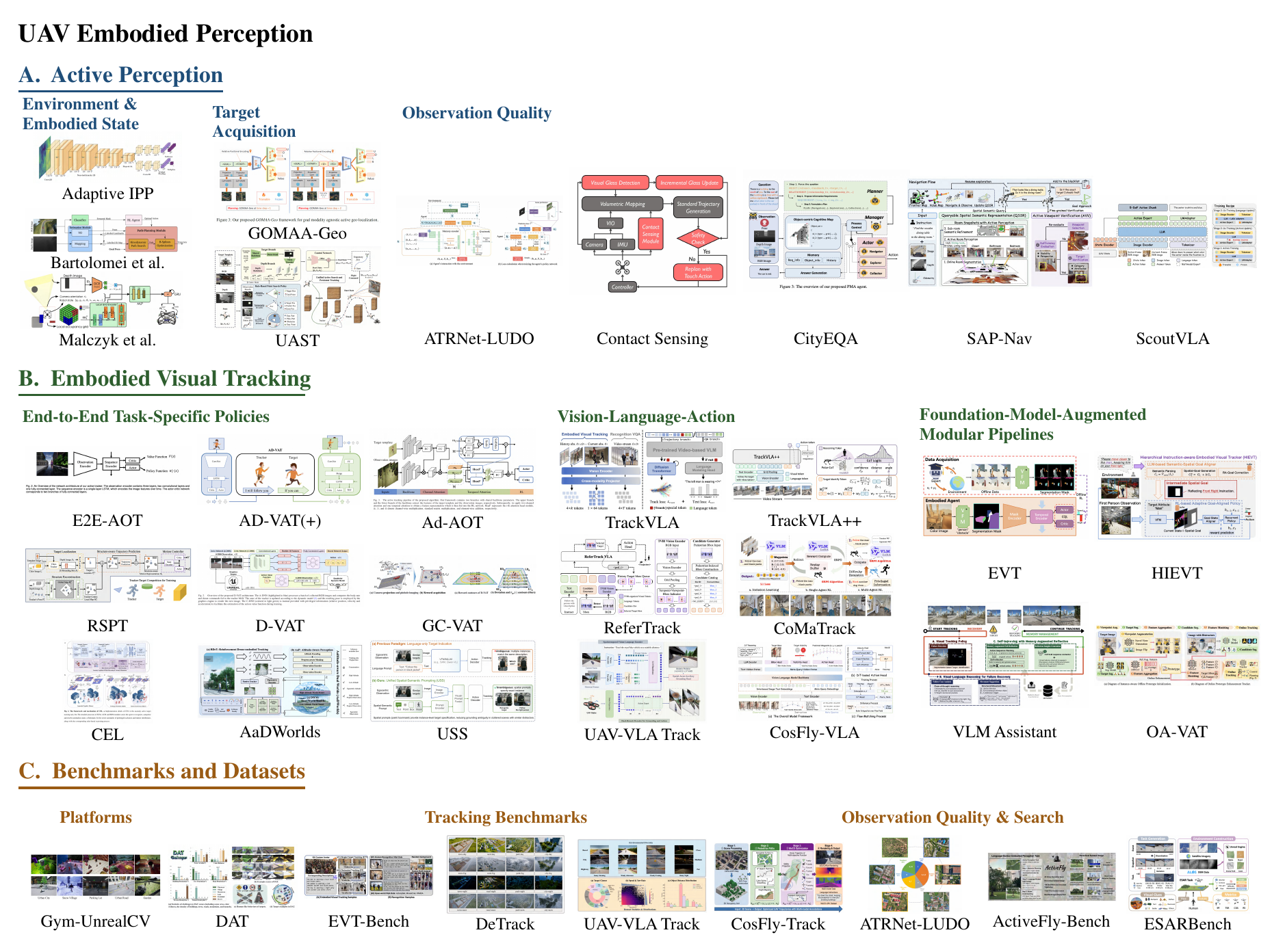}
\caption{Overview of UAV embodied perception.
(A) Active perception includes sensing for the environment and embodied state, represented by Adaptive IPP \cite{ruckin2022adaptive}, Bartolomei et al. \cite{bartolomei2021semantic}, and Malczyk et al. \cite{malczyk2026reinforcement}; target acquisition, represented by GOMAA-Geo \cite{sarkar2024gomaa} and UAST \cite{qin2026uast}; and observation quality, represented by ATRNet-LUDO \cite{liu2026toward}, Contact Sensing \cite{chen2025active}, CityEQA \cite{zhao2025cityeqa}, SAP-Nav \cite{pei2026sap}, and ScoutVLA \cite{lu2026scoutvla}.
(B) Embodied visual tracking includes end-to-end task-specific policies, including E2E-AOT \cite{luo2018end}\cite{luo2019end}, AD-VAT \cite{zhong2019advat}, AD-VAT+ \cite{zhong2019advatplus}, Ad-AOT \cite{xi2021anti}, RSPT \cite{zhong2023rspt}, D-VAT \cite{dionigi2024d}, GC-VAT \cite{sun2026open}, CEL \cite{wu2025cognitive}, AaDWorlds \cite{hu2026detrack}, and USS \cite{xie2026uss}; VLA methods, including TrackVLA \cite{wang2025trackvla}, TrackVLA++ \cite{liu2025trackvlapp}, ReferTrack \cite{ye2026refertrack}, CoMaTrack \cite{liu2026comatrack}, UAV-Track VLA \cite{zhang2026uavtrack}, and CosFly-VLA \cite{ren2026cosfly}; and foundation-model-augmented modular pipelines, including EVT \cite{zhong2024empowering}, HIEVT \cite{wu2025hierarchical}, VLM Assistant \cite{wu2025vlm}, and OA-VAT \cite{sun2026instance}.
(C) Representative benchmarks and datasets include Gym-UnrealCV \cite{zhong2017gym}\cite{qiu2017unrealcv}, DAT \cite{sun2026open}, EVT-Bench \cite{wang2025trackvla}, DeTrack \cite{hu2026detrack}, UAV-VLA Track \cite{zhang2026uavtrack}, CosFly-Track \cite{wang2026cosfly}, ATRNet-LUDO \cite{liu2026toward}, ActiveFly-Bench \cite{zhang2026activefly}, and ESARBench \cite{zhang2026esarbench}.}
\label{fig:perception}
\end{figure*}

\section{UAV Embodied Perception}

Embodied perception couples information acquisition with physical action. Unlike passive perception, where observations are treated as fixed inputs, UAVs can actively change their position, viewpoint, sensor configuration, or interaction strategy to obtain missing information. As illustrated in Fig.~\ref{fig:perception}, we organize UAV embodied perception into active perception and embodied visual tracking. Active perception focuses on selecting informative sensing actions, while embodied visual tracking integrates perception with continuous UAV motion to maintain informative observations of dynamic targets.

\subsection{Active Perception}

Active perception enables UAVs to acquire task-relevant information by selecting appropriate sensing actions. Existing methods mainly consider three objectives: improving environmental and state understanding, locating specified targets, and obtaining observations with sufficient quality for subsequent reasoning.

\textbf{Environment and Embodied State.}
When information is distributed across the environment, UAV motion directly affects sensing coverage and observation quality. Adaptive informative path planning (IPP) combines Monte Carlo tree search with a learned policy-value network to reduce sensing-action evaluation cost and support transfer from simulation to real farmland \cite{ruckin2022adaptive}. Semantic-aware planning further incorporates localization reliability into the planning objective, allowing UAVs to avoid observations that degrade visual odometry \cite{bartolomei2021semantic}. Malczyk et al. jointly optimize camera orientation and body motion to improve environment coverage during goal-directed flight \cite{malczyk2026reinforcement}.

\textbf{Target Acquisition.}
Target acquisition requires UAVs to search for initially unknown target locations under limited observations and motion constraints. GOMAA-Geo aligns text, ground imagery, and aerial imagery within a shared representation and uses observation history to guide search \cite{sarkar2024gomaa}. UAST extends target acquisition by integrating search, tracking, and re-acquisition through a mapping-free color-depth framework \cite{qin2026uast}.

\textbf{Observation Quality Regulation.}
Active perception can also regulate viewpoint, distance, and sensor configuration to obtain more informative observations. ATRNet-LUDO selects viewpoints to reduce recognition failures caused by unfavorable aerial views \cite{liu2026toward}, while contact sensing complements vision when remote observations are insufficient \cite{chen2025active}. For language-conditioned tasks, CityEQA separates navigation and viewpoint refinement for embodied question answering \cite{zhao2025cityeqa}, SAP-Nav introduces view-quality-driven repositioning \cite{pei2026sap}, and ScoutVLA jointly reasons about missing evidence and generates viewpoint adjustment trajectories \cite{lu2026scoutvla}.

\subsection{Embodied Visual Tracking}
\label{sec:vat}

Embodied visual tracking requires UAVs to continuously regulate their own motion to maintain informative observations of moving targets. Compared with passive tracking, it couples target identification, temporal estimation, motion prediction, and control within a closed perception-action loop. Existing approaches can be categorized into task-specific policies, vision-language-action (VLA) methods, and foundation-model-augmented modular pipelines.

\textbf{Task-Specific Policies.}
Task-specific methods directly map observations to tracking actions using models designed for active tracking. Early studies established direct visuomotor tracking and explored adversarial target motion, geometric reasoning, motion prediction, and flexible target prompts \cite{luo2018end,luo2019end,zhong2019advat,zhong2019advatplus,zhong2023rspt,xie2026uss}. UAV-specific methods further incorporate aerial geometry and flight dynamics. Ad-AOT uses spatial-temporal attention to handle distractors and appearance changes \cite{xi2021anti}. D-VAT directly predicts thrust and body rates from monocular image sequences \cite{dionigi2024d}, while GC-VAT designs image-space rewards for aerial tracking control \cite{sun2026open}. CEL introduces recovery behaviors under severe occlusion \cite{wu2025cognitive}, and AaDWorlds uses altitude-aware world models to reason about resolution, field of view, occlusion, and collision risk \cite{hu2026detrack}.

\textbf{VLA Methods.}
VLA methods incorporate pretrained multimodal representations to jointly interpret target descriptions and generate actions. Ground-based approaches improve target recognition and control through unified representations, spatial reasoning, target memory, and interactive learning \cite{wang2025trackvla,liu2025trackvlapp,ye2026refertrack,liu2026comatrack}. UAV-Track VLA extends this paradigm to aerial tracking by compressing historical observations and generating continuous flight actions \cite{zhang2026uavtrack}. CosFly-VLA further considers long-duration occlusion recovery by jointly predicting target visibility, location, and flight actions \cite{ren2026cosfly}.

\textbf{Foundation-Model-Augmented Modular Pipelines.}
Modular approaches maintain explicit interfaces among perception, reasoning, planning, and control, using foundation models mainly for transferable representations while retaining dedicated planners or controllers for execution. Ground-based studies explore text-conditioned target representation and failure recovery \cite{zhong2024empowering,wu2025hierarchical,wu2025vlm}. OA-VAT extends this framework to UAVs by maintaining online target prototypes and using diffusion-based planning for obstacle-aware recovery under occlusion \cite{sun2026instance}.

\subsection{Benchmarks}

Evaluation of UAV embodied perception requires assessing not only perception accuracy but also whether sensing can be actively improved through motion and viewpoint regulation. Existing resources mainly include simulation platforms, task-specific benchmarks, and offline trajectory datasets. Simulation platforms provide controllable environments for embodied perception evaluation. Gym-UnrealCV supports active visual tracking in simulation environments \cite{zhong2017gym,qiu2017unrealcv}, while CARLA-Air extends evaluation toward urban and air-ground scenarios \cite{zeng2026carla}. Task-specific benchmarks evaluate different capabilities within the perception-action loop. Tracking benchmarks progressively introduce language-conditioned targets, aerial dynamics, and UAV-specific actions \cite{wang2025trackvla,hu2026detrack,zhang2026uavtrack,sun2026open}, while observation and search benchmarks focus on viewpoint regulation and active target acquisition \cite{liu2026toward,zhang2026activefly,zhang2026esarbench}. Offline trajectory datasets such as CosFly-Track provide fixed observation-action sequences for policy learning and trajectory prediction \cite{wang2026cosfly}, but do not directly evaluate closed-loop interaction.

%% file: sections/uav-world-models.tex
\section{UAV World Models}
\label{sec:world_models}

For UAVs, generating actions directly from raw visual observations often leads to reactive, short-sighted behaviors that fail in complex three-dimensional spaces. World models and multimodal foundation models endow aerial vehicles with the embodied capability to predict the spatiotemporal evolution of their environment, simulate the consequences of their own flight dynamics, and reason over complex tasks. By decoupling environment understanding from action generation, these models provide predictive priors and cognitive maps that mitigate sensor latency, bridge the sim-to-real gap, and guide long-horizon navigation. This section organizes their application in UAV intelligence through five core algorithmic affordances: spatial-temporal intelligence for MLLMs, predictive spatial representations, generative simulation, model-based flight control, and world-action modeling with vision-language integration.

\subsection{Spatial-Temporal Intelligence for MLLMs}
While the broader scope of this section focuses on predictive world models, establishing robust spatial-temporal intelligence is a fundamental prerequisite for any embodied aerial agent. Consequently, this subsection specifically reviews literature on general Large Language Models (LLMs) and Vision-Language Models (VLMs) that enhance drone spatial awareness and reasoning, even if they do not explicitly employ generative world modeling architectures. Despite demonstrating strong reasoning capabilities in general domains, MLLMs exhibit severe performance degradation when directly deployed in low-altitude UAV scenarios. This deficiency stems from an inherent ground-centric bias and a lack of explicit temporal-spatial awareness, leading to failures in fine-grained visual grounding, ego-motion perception, and cross-scale navigation \cite{dai2025mm, zhan2026uavbench}. Comprehensive benchmarks such as MM-UAVBench \cite{dai2025mm} and UAV-DualCog \cite{liu2026knowing} reveal that although general MLLMs can occasionally guess correct answers in visual question answering (VQA) tasks, they struggle to accurately ground objects with bounding boxes or track dynamic multi-agent interactions across temporal intervals.

To bridge this domain gap, recent efforts focus on providing large-scale, geometry-aware, and instruction-tuned datasets. Environments like AeroVerse \cite{yao2024aeroverse} and AirZoo \cite{cheng2026airzoo} generate massive image-text-pose alignments and dense metric depth annotations in high-fidelity three-dimensional (3D) simulations, explicitly forcing models to learn ego-centric spatial awareness and overcome severe perspective distortions. By fine-tuning on specialized datasets such as UAVIT-1M \cite{zhan2026uavbench}, MLLMs can significantly improve their regional detection and low-altitude VQA capabilities. At the architectural level, enhancing temporal intelligence requires explicit motion modeling; for instance, the SIS-Motion framework introduces an optical-flow-based motion encoder parallel to the visual stream, enabling the model to jointly reason about external spatial layout and the drone's own continuous kinematics \cite{zou2026self}. For complex Embodied Question Answering (EQA) in expansive urban spaces, hierarchical agents like PMA (Planner-Manager-Actor) construct object-centric cognitive maps to balance long-horizon path planning with fine-grained motion control, effectively mitigating spatial ambiguity and reducing navigation errors \cite{zhao2025cityeqa}.

\subsection{Predictive Spatial Representations}
Predicting raw pixel sequences is computationally inefficient for agile flight and fails to provide explicit geometric boundaries. To establish physical spatial awareness, Mapping-Aware Dreamer (MAD) substitutes pixel reconstruction with Occupancy and Visibility Grid Maps, forcing the latent state to encode local 3D geometry and decoupling visible areas from unmapped blind spots \cite{zhang2026mad}. When dealing with long-horizon visual generation, the Aerial World Model (ANWM) introduces a Future Frame Projection mechanism, injecting coarse 3D geometric priors into conditional diffusion to stabilize spatial layout generation \cite{zhang2025aerial}. Furthermore, to address semantic drift over long execution horizons, WorldVLN incorporates a closed-loop autoregressive backbone that periodically truncates imagined latents with real visual observations, rigidly anchoring the generative process to actual physical evolution \cite{zhao2026worldvln}.

\subsection{Generative Simulation and Data Synthesis}
The acquisition of real-world, high-dynamic aerial data is severely constrained by physical risks and airspace regulations. World models offer a scalable pathway to synthesize physically feasible interaction data and benchmark robustness. MotionScape highlights the limitations of current generation capabilities by providing a large-scale real-world dataset characterized by extreme six degrees of freedom (6-DoF) dynamics, exposing the temporal degradation vulnerabilities of existing generalist models \cite{guo2026motionscape}. To bypass manual 3D modeling pipelines, FlyMirage leverages LLMs alongside 3D Gaussian Splatting (3DGS) to automatically synthesize boundless, photorealistic, and collision-free navigable environments \cite{li2026flymirage}. Recognizing the impact of synthetic variability, systematic evaluations demonstrate that the capacity of self-supervised world models to generalize across environment randomization directly correlates with their success rate in subsequent physical real-world deployment \cite{zanatta2026generalization}.

\subsection{Model-Based Reinforcement Learning for Flight Control}
Model-based reinforcement learning (MBRL) allows UAVs to optimize highly agile policies entirely within latent imagination, circumventing sample-inefficient interactions in the physical world. Dream to Fly directly learns end-to-end continuous commands from raw pixels, fostering emergent perception-aware behaviors where the UAV autonomously points its camera at salient features to assist localization without explicit reward shaping \cite{romero2025dream}. To overcome the susceptibility of end-to-end policies to local optima, AirDreamer utilizes spatial memory from its world model to enable sparse-reward navigation. Combined with extensive domain randomization, this framework achieves zero-tuning sim-to-real transfer, autonomously discovering complex escape maneuvers such as reversing out of unmapped physical traps \cite{liu2026airdreamer}.

\subsection{World-Action Models and Vision-Language-Action Systems}
Integrating world models directly with action generation, and extending this to VLA frameworks, resolves the disconnect between semantic commands, environment prediction, and physically executable flight trajectories. For real-time onboard navigation, FlowPilot introduces a compact dual-stream mixture-of-transformers that jointly denoises future depth observations and executable trajectories via flow matching \cite{wang2026flowpilot}. By parameterizing actions as state-constrained Bernstein polynomials and running action-centric inference without decoding future video at test time, the model generates $C^2$-continuous, controller-trackable references within an 18 ms latency on edge hardware \cite{wang2026flowpilot}. Scaling similar coupled architectures to semantic tasks, WorldFly proposes a dual-branch model that aligns future state imagination with action prediction through flow matching, enabling the UAV to anticipate dramatic viewpoint shifts during complex urban canyon traversal \cite{zheng2026worldfly}. Similarly, ImagineUAV transforms high-level language instructions into latent video diffusion sequences, serving as a semantic bridge. A dedicated visual odometry extractor translates these visual hallucinations into 6-DoF waypoints, which are subsequently refined by a kinodynamic planner to guarantee physical safety and actuation feasibility before execution \cite{liu2026imagineuav}.

\begin{table*}[!t]
\centering
\caption{Representative World Models for UAV Embodied Intelligence}
\label{tab:world_models_summary}
\resizebox{\textwidth}{!}{%
\scriptsize
\setlength{\tabcolsep}{3.2pt}
\renewcommand{\arraystretch}{1.15}
\begin{tabular}{p{2.0cm}p{1.6cm}p{1.8cm}p{1.8cm}p{1.9cm}p{1.7cm}p{1.7cm}p{1.4cm}p{1.5cm}}
\toprule
\textbf{System} & \textbf{Target Task} & \textbf{Architecture} & \textbf{State Rep.} & \textbf{Env. Source} & \textbf{Action Space} & \textbf{Eval. Metrics} & \textbf{Deployment} & \textbf{Role} \\
\midrule
\multicolumn{9}{l}{\textit{Predictive Spatial Representations}} \\
MAD \cite{zhang2026mad} & Agile Nav & RSSM & OGM / VGM & DiffAero / Gazebo & Cont.\ ctrl. & SR, Spd & 0-Shot Real & Spat.\ Rep. \\
ANWM \cite{zhang2025aerial} & Visual Nav & Cond.\ DiT & RGB + FFP & UE5 & 4-DoF Pose & FID, SR & 0-Shot Real & Spat.\ Rep. \\
WorldVLN \cite{zhao2026worldvln} & VLN & Auto-Reg. & Trunc.\ Latent & Real Flight & 6-DoF Waypt. & SR, NDTW & Real Deploy & Spat.\ Rep. \\
\midrule
\multicolumn{9}{l}{\textit{Generative Simulation and Data Synthesis}} \\
MotionScape \cite{guo2026motionscape} & Data Synth. & Video Model & RGB + Text & Real Capture & 6-DoF Pose & FVD, Flow & N/A & Gen.\ Sim. \\
FlyMirage \cite{li2026flymirage} & Scene Gen. & LLM + 3DGS & 3DGS & Synth.\ $\rightarrow$ Sim & 6-DoF Traj. & SR, SPL, NE & Sim$\rightarrow$Real & Gen.\ Sim. \\
Zanatta et al.~\cite{zanatta2026generalization} & Gen.\ Eval. & RSSM & Latent (Depth) & AerialGym & Cont.\ ctrl. & Loss, SR & 0-Shot Real & Gen.\ Sim. \\
\midrule
\multicolumn{9}{l}{\textit{Model-Based Reinforcement Learning for Flight Control}} \\
Dream to Fly \cite{romero2025dream} & Agile Racing & RSSM & Latent & Flightmare & CTBR & Reward, Spd & HIL & MBRL \\
AirDreamer \cite{liu2026airdreamer} & Generalist Nav & RSSM & Latent (Depth) & OmniDrones & 6-DoF Cont. & SR & 0-Shot Real & MBRL \\
\midrule
\multicolumn{9}{l}{\textit{World-Action Models and Vision-Language-Action Systems}} \\
FlowPilot \cite{wang2026flowpilot} & Agile Nav & MoT & Depth + State & Sim + Real & Bernst.\ Poly. & SR, Spd, Lat. & Real Deploy & WAM \\
WorldFly \cite{zheng2026worldfly} & VLA Nav & Flow Match & Latent + Action & AirSim & 6-DoF Cont. & SR, SPL, NE & Sim Only & VLA Int. \\
ImagineUAV \cite{liu2026imagineuav} & VLN & Video Diff. & RGB $\rightarrow$ Token & UE5 / Real & 6-DoF Waypt. & SR, Delay & Real Deploy & VLA Int. \\
\bottomrule
\end{tabular}%
}
\parbox{\textwidth}{\vspace{2pt}\footnotesize\textit{Notes:} Role abbreviations: Spat.\ Rep., Gen.\ Sim., MBRL, VLA Int., and WAM. Architecture and representation abbreviations: RSSM (Recurrent State Space Model), DiT, Auto-Reg., Cond., OGM/VGM, FFP, 3DGS, and MoT. Action abbreviations: Cont.\ ctrl. (Continuous control), Waypt., Traj., Bernst.\ Poly., and CTBR (Collective Thrust and Body Rates). Deployment labels: HIL (Hardware-in-the-Loop), 0-Shot Real (Zero-shot real-world deployment without fine-tuning), and Real Deploy (Trained partly on real data and deployed onboard). Metric abbreviations: SR, Spd, NE, and Lat.; FID/FVD (Fréchet Inception/Video Distance), SPL (Success weighted by Path Length), and NDTW (Normalized Dynamic Time Warping).}
\end{table*}

%% file: sections/uav-embodied-planning.tex
\section{UAV Embodied Planning}
\label{sec:uav-embodied-planning}

Embodied planning transforms observations, task conditions, and platform states into actions or executable references through continuous interaction with the environment. For UAVs, planning must operate under changing observations, incomplete sensing, fast vehicle dynamics, limited onboard computation, and execution feedback. It therefore provides a key interface between high-level task intent and physically executable aerial behavior.

Planning is involved in multiple embodied capabilities throughout this survey. Active perception requires viewpoint and search planning; navigation involves route, subgoal, exploration, and motion planning; manipulation requires approach and interaction planning; and collaboration introduces task allocation and coordinated decision making. Because these capabilities have their own task formulations and methodological developments, their task-specific planning components are reviewed in the corresponding sections to avoid duplication.

This section therefore focuses on two complementary perspectives. We first review task-independent \emph{planning paradigms}, including explicit model-based, policy-centric, structured hybrid, world-model, foundation-model, agentic, and role-specialized multi-agent planning. We then examine representative tasks that are directly coupled with UAV flight, including racing, obstacle avoidance, and intent-guided flight, to illustrate how different planning approaches interact with latency, partial observability, safety, semantics, vehicle dynamics, and physical executability.


\subsection{Planning Paradigms}
\label{subsec}

Planning paradigms differ primarily in where planning competence resides and how observations, task conditions, and physical constraints are transformed into executable decisions. The paradigms reviewed below are not mutually exclusive. Practical UAV systems often combine several of them, particularly when semantic reasoning and long-horizon decision making must coexist with real-time geometric and dynamic constraints.

\subsubsection{Explicit Model-Based Planning}

Explicit model-based planning represents geometry, dynamics, objectives, and constraints directly within search or optimization. Classical approaches construct dynamically feasible trajectories from prescribed goals or waypoints, while subsequent methods improve replanning efficiency, obstacle avoidance, and perception awareness~\cite{mellinger2011minimum,zhou2019robust,zhou2020ego,tordesillas2019faster,zhou2021raptor,chen2024apace}. Minimum-time formulations further push explicit optimization toward the dynamic limits of aerial platforms~\cite{romero2022time,teissing2024real}. In these methods, collision geometry, trajectory continuity, actuator limits, and related feasibility conditions remain explicitly represented, while observations and task intent are first converted into structured states, maps, goals, or costs.

\subsubsection{Policy-Centric Planning}

Policy-centric planning moves part of the planning computation from online search or optimization into a learned task-specific policy. Instead of repeatedly solving a planning problem during execution, the policy learns a mapping from observations or states to decisions, waypoints, trajectories, velocities, or control commands~\cite{song2021autonomous,ou2021autonomous,kim2025rapid,wu2026precise}. Planning structure is therefore encoded through demonstrations, privileged supervision, reward functions, or training distributions. Recurrent policies incorporate observation history~\cite{ou2021autonomous}, teacher-student learning transfers privileged planning information~\cite{song2023learning}, and safety-oriented reinforcement learning introduces task constraints during policy optimization~\cite{xiao2026time}.

\subsubsection{Structured Hybrid Planning}

Structured hybrid planning distributes planning responsibility between learned components and explicitly modeled modules. Learned models are typically used where perception, prediction, semantic reasoning, or motion generation is difficult to specify analytically, while geometric planning, trajectory execution, or safety mechanisms retain explicit physical structure.

Learned perception can produce compact spatial references for downstream planning~\cite{kaufmann2018deep}; expert behavior can be distilled into fast trajectory generation while retaining a trajectory-level execution interface~\cite{loquercio2021learning,tordesillas2023deep}; and learned policies can be combined with explicit safety mechanisms~\cite{zhang2026high}. The same organization appears when foundation models provide semantic guidance that is subsequently processed through geometric search, waypoint refinement, or dynamics-aware execution~\cite{ye2025vlm,xiao2025fm}.

\subsubsection{World-Model Planning}

World-model planning introduces an internal predictive representation of how states or observations may evolve under future actions. Whereas policy-centric approaches primarily learn what action to take from the current or recent state, world models additionally provide a basis for reasoning about possible future outcomes.

Aerial world models have been developed for latent dynamics, spatial memory, future observation prediction, and action-conditioned decision making~\cite{romero2025dream,zhang2025aerial,zhang2026mad,wang2026flowpilot}. Some approaches emphasize compact spatial representations that preserve planning-relevant structure, while others use generative prediction to imagine future observations or trajectories. More recent systems combine predictive models with VLA generation or kinodynamic planning~\cite{zheng2026worldfly,liu2026imagineuav,zhao2026worldvln}.

\subsubsection{Foundation-Model Planning}

Foundation-model planning extends planning conditions from predefined geometric goals toward open-ended visual and linguistic intent. Vision-language and VLA models allow planning to incorporate objects, landmarks, attributes, instructions, and high-level task descriptions that are difficult to express through conventional geometric variables~\cite{hu2025see,sautenkov2025uav,chen2026aerialvla,sun2026autofly}.

Foundation models can participate at different levels of the planning pipeline. Some systems predict actions or motion references directly, while others generate semantic goals, spatial relations, waypoints, or planning priors that are subsequently processed by geometric or dynamics-aware modules~\cite{li2025skyvln,ye2025vlm,sanyal2024asma,xiao2025fm}.

\subsubsection{Agentic Planning}

Agentic planning changes planning from a single inference into an iterative decision process. An agentic planner can decompose a long-horizon objective, maintain memory, acquire additional information, evaluate task progress, and revise its decisions as observations or task conditions change.

Representative systems introduce combinations of semantic decomposition, structured memory, active search, verification, recovery, and replanning into UAV embodied tasks~\cite{zhang2025citynavagent,shao2026finecog,qi2026parse,zhang2026apex,zhou2026memory}. Related mechanisms also appear in embodied question answering, search and rescue, and active tracking~\cite{zhao2025cityeqa,zhang2026esarbench,liu2025trackvlapp,wu2025vlm}.

\subsubsection{Role-Specialized Multi-Agent Planning}

Role-specialized multi-agent planning distributes the internal planning process across multiple cognitive agents with complementary responsibilities. Here, multi-agent refers to multiple software or foundation-model agents within a unified planning system rather than multiple physical UAVs. The resulting system still acts as a single planner from the perspective of the embodied platform.

Different agents can assume specialized roles such as task decomposition, state and perception analysis, plan generation, feasibility checking, verification, and reflection. Their interaction may follow a sequential pipeline, where intermediate results are passed from one role to another, or an iterative process in which critics and verifiers return detected inconsistencies to planning agents for revision. Different roles may also use distinct prompts, tools, memories, or foundation models according to their reasoning requirements.

\subsection{Task-Oriented Review}
\label{subsec:planning-task-review}

Planning appears throughout perception, navigation, manipulation, and collaboration, whose task-specific formulations are reviewed in their corresponding sections. Here, we instead examine three representative flight-coupled settings that expose distinct planning requirements: Dynamic-Limit Flight, Collision-Constrained Flight, and Intent-Conditioned Flight. The first stresses latency and vehicle dynamics near the flight envelope, the second emphasizes safe decision-making under incomplete environmental observations, and the third examines how geometric or semantic intent is converted into physically executable motion.

Table~\ref{tab:embodied_planning_comparison} compares representative methods according to planner input, planning paradigm, action abstraction, feasibility handling, and validation. Together, these dimensions reveal how different planning structures connect task conditions and observations to executable aerial behavior.

\begin{table*}[!t]
\centering
\caption{Representative planning methods for flight-coupled UAV tasks.}
\label{tab:embodied_planning_comparison}

\scriptsize
\setlength{\tabcolsep}{3.2pt}
\renewcommand{\arraystretch}{1.15}

\resizebox{\textwidth}{!}{%
\begin{tabular}{
p{3.0cm}
p{2.2cm}
p{2.4cm}
p{2.0cm}
p{3.3cm}
p{1.2cm}
}
\toprule
\textbf{Method} &
\textbf{Planner Input} &
\textbf{Planning Paradigm} &
\textbf{Action Abstraction} &
\textbf{Feasibility Handling} &
\textbf{Validation} \\
\midrule

\multicolumn{6}{l}{Dynamic-Limit Flight} \\
\addlinespace[1.5pt]

Deep Drone Racing \cite{kaufmann2018deep} &
RGB &
Structured Hybrid &
Waypoint &
Downstream planning and control &
Real Flight \\

Autonomous Drone Racing \cite{song2021autonomous} &
State &
Policy-Centric &
Control &
Implicit in learned policy &
Simulation \\

Minimum-Time Flight \cite{penicka2022learning} &
State &
Explicit Model-Based &
Trajectory &
Explicit optimization &
Simulation \\

Perception-Aware Flight \cite{song2023learning} &
Depth &
Policy-Centric &
Control &
Perception-aware objective &
HIL \\

Swift \cite{kaufmann2023champion} &
Visual-inertial state &
Structured Hybrid &
Control &
Estimation and execution stack &
Real Flight \\

Dream to Fly \cite{romero2025dream} &
Image sequence &
World-Model &
Control &
Learned predictive dynamics &
HIL \\

Time-Optimal Safe RL \cite{xiao2026time} &
State &
Policy-Centric &
Control &
Reward shaping &
Simulation \\

\midrule
\multicolumn{6}{l}{Collision-Constrained Flight} \\
\addlinespace[1.5pt]

D3RQN \cite{ou2021autonomous} &
Depth history &
Policy-Centric &
Discrete decision &
Reward shaping &
Simulation \\

Guidance Exploration \cite{roghair2021vision} &
RGB &
Policy-Centric &
Discrete decision &
Reward shaping &
Simulation \\

High-Speed Flight \cite{loquercio2021learning} &
Depth + state &
Structured Hybrid &
Trajectory &
Trajectory feasibility &
Real Flight \\

Deep-PANTHER \cite{tordesillas2023deep} &
State + prediction &
Structured Hybrid &
Trajectory &
Trajectory feasibility &
Simulation \\

RAPID \cite{kim2025rapid} &
RGB &
Policy-Centric &
Waypoint &
Downstream controller &
Real Flight \\

Vision-Guided Outdoor Flight \cite{dutta2025vision} &
Depth history &
Policy-Centric &
Velocity &
Downstream flight stack &
Real Flight \\

Fly360 \cite{zhang2026fly360} &
Panoramic RGB history &
Policy-Centric &
Velocity &
Implicit in learned policy &
Real Flight \\

Safety-Shielded RL \cite{zhang2026high} &
Depth &
Structured Hybrid &
Control &
HOCBF safety filter &
Real Flight \\

Sensorimotor Gap Policy \cite{wu2026precise} &
RGB &
Policy-Centric &
Control &
Task-specific geometry &
Real Flight \\

\midrule
\multicolumn{6}{l}{Intent-Conditioned Flight} \\
\addlinespace[1.5pt]

SketchPlan \cite{norelius2026sketchplan} &
Sketch &
Structured Hybrid &
Path &
Collision checking and refinement &
Deployment \\

VLM-RRT \cite{ye2025vlm} &
Vision + language &
Structured Hybrid &
Sampling bias &
Geometric collision checking &
Simulation \\

FM-Planner \cite{xiao2025fm} &
Vision + language &
Foundation-Model &
Waypoint &
Waypoint refinement &
MoCap \\

\bottomrule
\end{tabular}%
}

\parbox{\textwidth}{
\vspace{2pt}
\footnotesize
Notes:
HIL denotes hardware-in-the-loop,
HOCBF denotes high-order control barrier function,
and MoCap denotes motion-capture-assisted flight.
}
\end{table*}

\subsubsection{Dynamic-Limit Flight}

Dynamic-limit flight examines planning when the UAV operates close to its motion and reaction limits, as in racing and minimum-time traversal. Increasing speed shortens the available sensing and planning horizon while making trajectory feasibility, perception quality, and control accuracy more tightly coupled.

Methods differ primarily in where this trade-off is resolved. Explicit optimization directly incorporates vehicle dynamics and minimum-time objectives \cite{penicka2022learning}, while policy-centric approaches encode aggressive behavior through learning objectives and demonstrations \cite{song2021autonomous,xiao2026time}. Other systems retain more structure between perception and execution: Deep Drone Racing predicts intermediate waypoints from images \cite{kaufmann2018deep}, Swift combines visual-inertial estimation with learned control \cite{kaufmann2023champion}, and Perception-Aware Flight transfers privileged planning knowledge into a depth-conditioned policy \cite{song2023learning}. Dream to Fly further introduces predictive latent dynamics from image histories \cite{romero2025dream}. Dynamic-limit flight therefore exposes a central trade-off between planning latency, explicit physical structure, and the robustness required near the UAV flight envelope.

\subsubsection{Collision-Constrained Flight}

Collision-constrained flight focuses on reaching or progressing toward a goal while maintaining safety in cluttered and only partially observed environments. Obstacles may be occluded, outside the current field of view, or moving, making planning dependent on both immediate geometry and temporal context.

Policy-centric methods encode this relationship directly from observation histories or local visual inputs. D3RQN uses recurrent reasoning over depth observations \cite{ou2021autonomous}, while Guidance Exploration improves policy learning under sparse environmental feedback \cite{roghair2021vision}. Later approaches expose more explicit motion structure through trajectories, waypoints, or velocities. High-Speed Flight and Deep-PANTHER generate dynamically feasible trajectories \cite{loquercio2021learning,tordesillas2023deep}, RAPID predicts waypoint sequences \cite{kim2025rapid}, and Vision-Guided Outdoor Flight and Fly360 exploit temporal or panoramic observations for local motion generation \cite{dutta2025vision,zhang2026fly360}.

Safety can be incorporated at different stages of this process. Reinforcement learning (RL) can encode collision avoidance within the optimization objective, while Safety-Shielded RL adds an explicit safety filter before physical execution \cite{zhang2026high}. Direct sensorimotor policies provide another solution for highly constrained maneuvers such as gap traversal \cite{wu2026precise}. These methods illustrate how collision-constrained planning balances perceptual coverage, reaction latency, motion representation, and explicit safety enforcement.

\subsubsection{Intent-Conditioned Flight}

Intent-conditioned flight introduces an external condition that specifies how or where the UAV should move. Unlike conventional planning from predefined goals, the condition may be a human sketch, visual reference, natural-language instruction, or semantic task description and must therefore be translated into a representation compatible with geometric and dynamic execution.

Structured conditions can be converted into explicit motion references. SketchPlan translates a human-provided sketch into a three-dimensional path and applies collision checking before execution \cite{norelius2026sketchplan}. Semantic conditions require an additional grounding stage between high-level intent and physical motion. VLM-RRT uses vision-language reasoning to bias geometric search while retaining explicit collision checking \cite{ye2025vlm}, whereas the foundation-model Planner (FM-Planner) derives spatial guidance from FM reasoning and subsequently refines the resulting waypoints before execution \cite{xiao2025fm}. These methods demonstrate a common planning pattern in which semantic reasoning determines where or how the UAV should move, while downstream geometric or dynamics-aware mechanisms ensure that the resulting motion remains executable.

%% file: sections/uav-vision-language-navigation.tex
\section{UAV Embodied Navigation}
\label{sec:uav-embodied-navigation}

Embodied navigation transforms human or task intent into goal-directed UAV motion through continuous interaction with the environment. Compared with ground navigation, aerial navigation operates in three-dimensional space with changing altitude, heading, and viewpoint, making perception, spatial grounding, memory, and motion tightly coupled. Navigation must therefore connect high-level objectives with physically executable flight under constraints on dynamics, sensing, computation, energy, and safety.

\begin{figure*}[t]
  \centering
  \includegraphics[width=\textwidth]{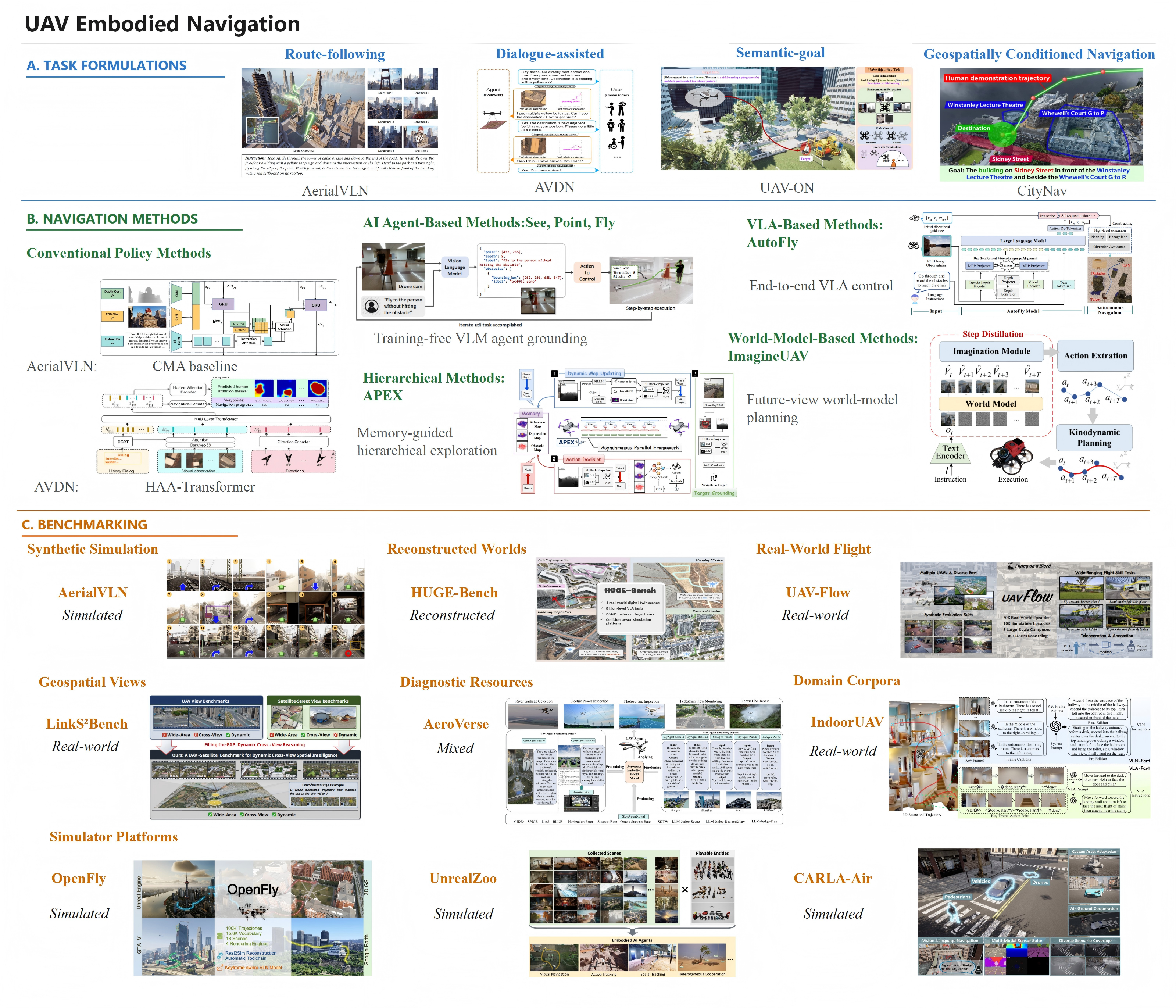}
  \caption{Overview of UAV embodied navigation.
  (A) Task formulations include route-following navigation, represented by AerialVLN \cite{liu2023aerialvln}; dialogue-assisted navigation, represented by AVDN \cite{fan2023aerial}; semantic-goal navigation, represented by UAV-ON \cite{xiao2025uav}; and geospatially conditioned navigation, represented by UAV-VLA \cite{sautenkov2025uav}.
  (B) Navigation methods include task-specific policies, represented by the CMA baseline of AerialVLN \cite{liu2023aerialvln} and HAA-Transformer of AVDN \cite{fan2023aerial}; hierarchical methods, represented by APEX \cite{zhang2026apex}; AI agent methods, represented by See, Point, Fly \cite{hu2025see}; VLA methods, represented by AutoFly \cite{sun2026autofly}; and world-model methods, represented by ImagineUAV \cite{liu2026imagineuav}.
  (C) Representative benchmarking resources include AerialVLN \cite{liu2023aerialvln}, HUGE-Bench \cite{guo2026huge}, UAV-Flow \cite{wang2026uav}, LinkS2Bench \cite{liu2026vlms}, AeroVerse \cite{yao2026aeroverse}, IndoorUAV \cite{liu2026indooruav}, OpenFly \cite{gao2026openfly}, UnrealZoo \cite{zhong2025unrealzoo}, and CARLA-Air \cite{zeng2026carla}.}
  \label{fig:nav-overview}
\end{figure*}

As summarized in Fig.~\ref{fig:nav-overview}, we review UAV embodied navigation from three perspectives. Task formulations distinguish how navigation intent is specified and grounded. Navigation methods are organized according to the dominant mechanism that converts grounded intent into motion. Benchmarks characterize how these capabilities are evaluated under different environments, action interfaces, and interaction regimes.

\subsection{Task Formulations}
\label{sec:nav-tasks}

UAV navigation tasks differ primarily in how the intended route, destination, or target is specified. We organize existing formulations into Route-Following Navigation, Dialogue-Assisted Navigation, Semantic-Goal Navigation, and Geospatially Conditioned Navigation. These formulations may coexist within a practical mission, but they impose different requirements on intent grounding, spatial reasoning, memory, exploration, and motion generation.

\subsubsection{Route-Following Navigation}

Route-following navigation provides an ordered language description of landmarks, spatial relations, viewpoints, and motion transitions. The UAV must maintain correspondence between instruction segments, observations, and navigation progress throughout the trajectory. This is particularly challenging in aerial settings because altitude, heading, and viewpoint changes can substantially alter landmark appearance, while local grounding errors may accumulate over long-distance flight.

AerialVLN establishes language-guided aerial route following in outdoor environments \cite{liu2023aerialvln}. Subsequent approaches improve long-horizon instruction grounding through historical context, landmark and direction reasoning, coarse-to-fine decisions, and recovery mechanisms \cite{ding2026history,ning2026lookasidevln,li2025skyvln}. CityNav and TravelUAV further extend this formulation toward city-scale navigation, geographic context, human demonstrations, and more realistic continuous flight \cite{lee2025citynav,wang2025towards}. The main progression is therefore from local instruction matching toward sustained instruction-motion alignment under changing aerial viewpoints.

\subsubsection{Dialogue-Assisted Navigation}

Dialogue-assisted navigation allows the UAV to refine incomplete or ambiguous task instructions through interaction. The key problem is not simply processing dialogue history, but using communication to reduce uncertainty about landmarks, destinations, spatial relations, or required behavior.

AVDN introduces multi-round interaction into aerial navigation by combining dialogue history with visual grounding and waypoint prediction \cite{fan2023aerial}. Later methods improve landmark grounding and structured spatial memory across dialogue turns \cite{su2025learning,qi2026parse}, while agent-based and VLA approaches further explore clarification, reasoning, and joint question-action decisions \cite{qi2026parse,chen2026aerialvla}. This formulation extends aerial navigation from passive instruction following toward interactive intent grounding.

\subsubsection{Semantic-Goal Navigation}

Semantic-goal navigation specifies what the UAV should find or reach without providing its location. The goal may correspond to an object, region, scene, or open-vocabulary concept, making active exploration a necessary part of navigation.

Aerial settings amplify this challenge because targets may appear small, become occluded, or change substantially with altitude and viewpoint, while exhaustive search is constrained by flight time and energy. UAV-ON establishes aerial semantic-goal navigation as a dedicated benchmark \cite{xiao2025uav}. Recent systems combine multimodal target grounding with spatial or semantic memory and learned exploration to improve search efficiency \cite{zhang2026apex,zhou2026memory}. The central problem is therefore the closed-loop integration of recognition, memory, exploration, and stopping rather than target recognition alone.

\subsubsection{Geospatially Conditioned Navigation}

Geospatially conditioned navigation specifies the destination or route through an external spatial reference such as satellite imagery, maps, geographic coordinates, or overhead views. The UAV must align this allocentric information with its egocentric observations before it can generate local motion.

This formulation is particularly relevant to UAVs because many aerial missions are naturally defined at geographic scale. UAV-VLA explores navigation conditioned on satellite imagery and language \cite{sautenkov2025uav}, while CityNav incorporates geographic and semantic context into city-scale navigation \cite{lee2025citynav}. LinkS2Bench isolates cross-view alignment between UAV observations and satellite imagery \cite{liu2026vlms}. Such grounding becomes embodied navigation when the inferred geographic relation is further translated into closed-loop flight behavior.

\subsection{Navigation Methods}

Navigation methods differ primarily in how grounded intent is transformed into executable motion. We organize representative approaches into Task-Specific Policies, Hierarchical Methods, AI Agent Methods, VLA Methods, and World-Model Methods. The categories describe dominant architectural mechanisms rather than mutually exclusive system designs.

\subsubsection{Task-Specific Policies}

Task-specific policies learn a dedicated mapping from observations and task conditions to navigation actions under a predefined task and action interface. Inputs may include images, language, dialogue, or platform state, while outputs range from discrete actions to waypoints and motion commands.

AerialVLN and AVDN establish this paradigm for route following and dialogue-assisted navigation \cite{liu2023aerialvln,fan2023aerial}, while later methods improve multimodal grounding, history modeling, and action refinement \cite{su2025learning,ding2026history}. Their strength lies in direct optimization for the target task, but generalization remains closely related to the vocabulary, observation model, action interface, and training environment.

\subsubsection{Hierarchical Methods}

Hierarchical methods separate navigation decisions across spatial or temporal scales. High-level components maintain task progress, semantic memory, or subgoals, while lower-level components generate local motion and respond to changing observations.

CityNavAgent, SkyVLN, and APEX use such organization for long-range navigation, recovery, and semantic search \cite{zhang2025citynavagent,li2025skyvln,zhang2026apex}. Hierarchy is particularly useful for UAVs because mission-level reasoning may span large spatial regions while flight execution must respond rapidly to local geometry and changing viewpoints.

\subsubsection{AI Agent Methods}

AI agent methods place pretrained language or multimodal models within an iterative reasoning loop that can use memory, spatial representations, and external tools. Rather than learning a complete navigation policy for a fixed benchmark, the agent interprets task intent and generates intermediate navigation decisions that are grounded into executable motion.

See, Point, Fly converts model reasoning into image-space waypoints for downstream geometric execution \cite{hu2025see}. Other systems use structured memory, landmark reasoning, and explicit task reasoning to support dialogue or long-range navigation \cite{qi2026parse,ning2026lookasidevln}. These methods benefit from broad semantic knowledge but depend strongly on reliable spatial grounding and execution interfaces.

\subsubsection{VLA Methods}

VLA methods integrate multimodal understanding and action generation within a learned model with an explicit action interface. Compared with AI agent methods that typically reason through external modules, VLA methods directly specialize pretrained multimodal representations for outputs such as waypoints, velocities, or flight commands.

Current UAV studies apply VLA models to dialogue-conditioned navigation, continuous flight, obstacle avoidance, and low-level action generation \cite{chen2026aerialvla,xu2026aerialvla,sun2026autofly,chen2025grad}. This tighter coupling reduces the separation between semantic understanding and execution, but generated actions must still satisfy flight dynamics, collision constraints, latency, and real-time control requirements.

\subsubsection{World-Model Methods}

World-model methods introduce predictive representations of future observations, states, or action consequences to support lookahead navigation. Instead of selecting actions only from current observations, they allow navigation decisions to account for anticipated future outcomes.

WorldFly, ImagineUAV, and WorldVLN use predictive representations or imagined trajectories to support navigation and long-horizon reasoning \cite{zheng2026worldfly,liu2026imagineuav,zhao2026worldvln}. FineCogNav and ANWM further connect instruction reasoning with future-state prediction \cite{shao2026finecog,zhang2025aerial}. Their potential lies in moving aerial navigation from reactive decision-making toward predictive behavior, although prediction quality remains sensitive to large viewpoint changes and environmental variation.

\subsection{Benchmarks}
\label{sec:nav-benchmarking}

UAV navigation resources serve three different roles: complete navigation benchmarks, supporting grounding or diagnostic benchmarks, and configurable simulation platforms. Navigation benchmarks evaluate closed-loop task execution, while grounding resources isolate capabilities such as geographic localization or embodied reasoning. Simulation platforms provide reusable environments in which different navigation tasks and action interfaces can be instantiated.

Representative benchmarks cover the four task formulations discussed above. AerialVLN, TravelUAV, and CityNav evaluate route following under increasingly realistic spatial and execution settings \cite{liu2023aerialvln,wang2025towards,lee2025citynav}. AVDN evaluates dialogue-assisted navigation \cite{fan2023aerial}, while UAV-ON focuses on semantic-goal search \cite{xiao2025uav}. HUGE-Bench extends evaluation toward multi-stage tasks involving higher-level multimodal reasoning and execution \cite{guo2026huge}. LinkS2Bench instead evaluates geospatial grounding rather than complete navigation \cite{liu2026vlms}.

A central distinction is whether evaluation is closed loop. In closed-loop navigation, each action changes the UAV state and therefore affects subsequent observations. Offline resources instead evaluate fixed observations or trajectories and do not directly test perception-action coupling. Configurable platforms such as UnrealZoo \cite{zhong2025unrealzoo}, CARLA-Air \cite{zeng2026carla}, and OpenFly \cite{gao2026openfly} can support different interaction regimes depending on the instantiated task.

Evaluation should therefore capture more than endpoint success. Common measures assess task completion, path efficiency, goal proximity, trajectory consistency, collision, and task progress. For UAV embodied navigation, however, the most informative evaluation combines successful intent grounding with efficient, safe, and closed-loop physical execution.

%% file: sections/uav-embodied-manipulation.tex
\subsection{Embodied Grasping and Aerial Manipulation}
\label{sec:embodied_grasping}

\begin{figure*}[!t]
    \centering
    \includegraphics[width=\textwidth]{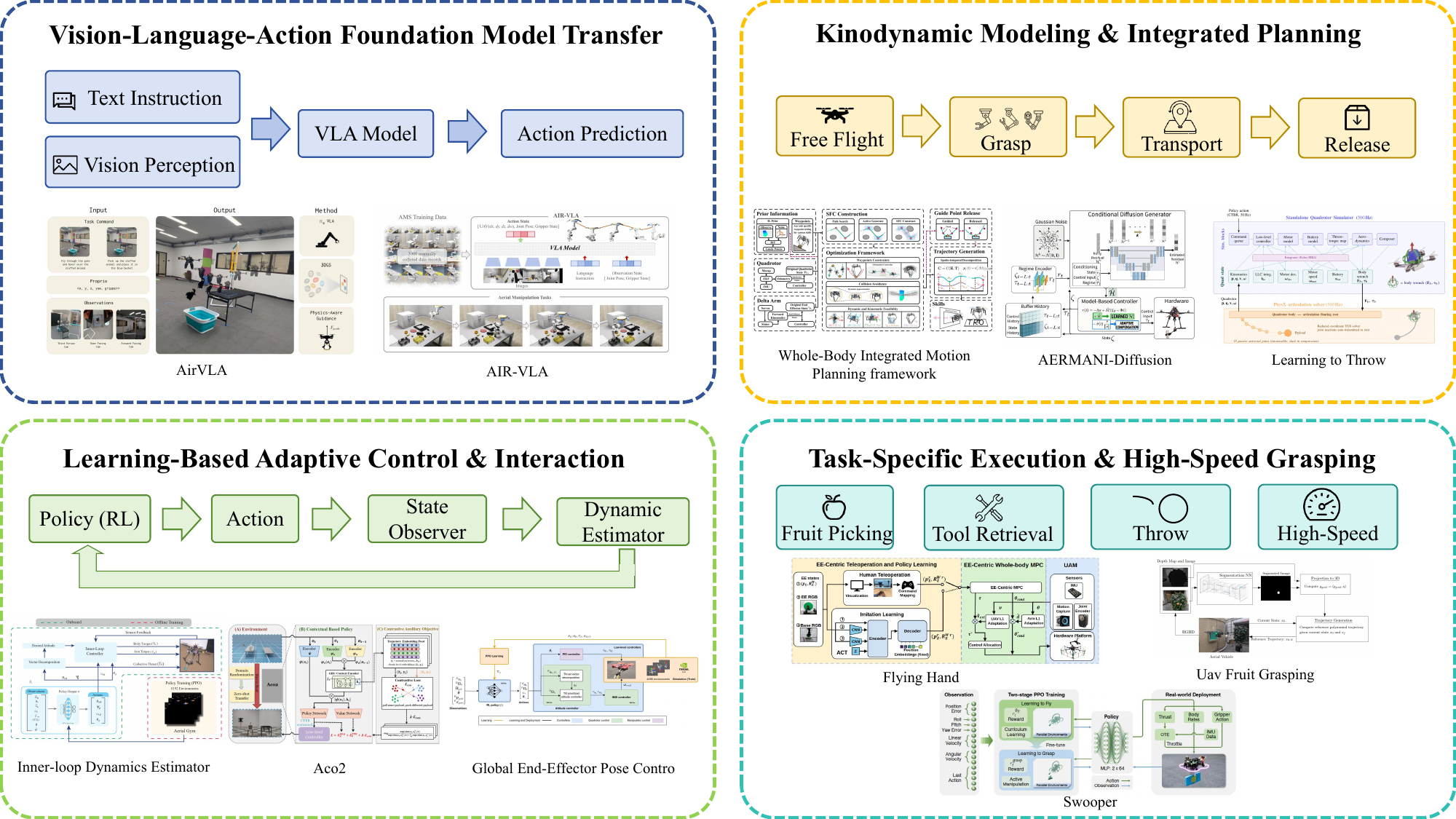}
    \caption{Overview of Embodied Grasping and Aerial Manipulation approaches across four components: (A) Vision-Language-Action Foundation Model Transfer, (B) Kinodynamic Modeling and Integrated Planning, (C) Learning-Based Adaptive Control and Interaction, and (D) Task-Specific Execution and High-Speed Grasping. Each category lists selected representative systems rather than an exhaustive set: foundation model transfer includes AIR-VLA and AirVLA \cite{sun2026air,tucker2026pi}; kinodynamic modeling and planning includes AERMANI-Diffusion, whole-body integrated motion planning frameworks, and learning to throw \cite{ujjawal2025aermani,deng2025whole,zhai2026learning}; learning-based control includes inner-loop dynamics estimators, Aco2, and global end-effector pose control \cite{singh2026reinforcement,jin2026autonomous,deshmukh2025global}; and task-specific execution includes UAV fruit grasping, Flying Hand, and Swooper \cite{baraban2021perception,he2025flying,huang2025swooper}.}
    \label{fig:aerial_manipulation}
\end{figure*}

Aerial grasping and manipulation fundamentally differ from fixed-base terrestrial robotics due to the inherent floating-base dynamics of the vehicle, the strong physical coupling between the UAV and the manipulator, and the continuous aerodynamic disturbances encountered during physical contact. To realize robust embodied grasping, intelligence architectures must concurrently address multi-stage interaction data scarcity, highly nonlinear kinodynamic constraints, and sim-to-real policy transfer gaps. 

\subsubsection{Data Regimes and Foundation Model Transfer}
High-fidelity interaction data is foundational for training end-to-end aerial manipulators. However, existing embodied datasets predominantly feature static ground bases, failing to capture the unique underactuated dynamics of UAVs in flight. To provide realistic floating-base interactions, the AIR-VLA benchmark introduces a comprehensive 12-DoF aerial manipulation simulation pipeline utilizing human-in-the-loop teleoperation \cite{sun2026air}. Evaluations on this benchmark reveal that while current physical foundation models can handle basic grasping, they suffer catastrophic performance degradation during long-horizon spatial navigation due to sequential error accumulation. Furthermore, transferring large-scale VLA models to flight platforms requires mitigating severe payload mass shifts, altitude droops, and inertia variations that violate the quasi-static assumptions of terrestrial algorithms. Systems like AirVLA overcome this by introducing payload-aware guidance during flow-matching inference and real-time temporal chunking, enabling the execution of composite navigate-then-grasp behaviors without computationally expensive retraining \cite{tucker2026pi}.

\subsubsection{Kinodynamic Modeling and Integrated Planning}
Executing aggressive physical interactions requires abandoning traditional methods that rigidly decouple the UAV base trajectory from the manipulator. In highly dynamic scenarios, unmodeled residual dynamics and regime-dependent inertia variations frequently render traditional analytical models invalid. To predict these non-stationary residual dynamics accurately, frameworks like AERMANI-Diffusion employ regime-conditioned diffusion models to capture the rapid transitions across free flight, payload grasping, and release phases \cite{ujjawal2025aermani}. At the motion planning level, whole-body integrated frameworks resolve high-dimensional kinodynamic conflicts by dynamically modeling the aerial system as a varying collision ellipsoid to safely traverse narrow spaces. By employing soft penalty mechanisms alongside imitation learning priors (e.g., Diffusion Transformers), these frameworks actively guide trajectory optimization away from poor local minima \cite{deng2025whole}. Furthermore, for extreme kinodynamic tasks such as throwing cable-suspended payloads, bridging the gap between offline trajectory optimization (TO) and online model predictive control (MPC) becomes imperative to handle the complex multi-body forces and frequent transitions between taut and slack cable states \cite{zhai2026learning}.

\subsubsection{Learning-Based Adaptive Control and Interaction}
The physical contact between a UAV manipulator and the environment introduces severe torque disturbances and center-of-mass variations. Deep Deterministic Policy Gradient (DDPG) methods have been widely adopted to decouple these complex internal interaction forces from the UAV chassis, ensuring robust attitude tracking under payload uncertainty \cite{liu2021ddpg}. To counter the instability caused by rapid mass and inertia changes during grasping, recent RL architectures incorporate inner-loop dynamics estimators that explicitly account for whole-body dynamic coupling \cite{singh2026reinforcement}. Dealing with heterogeneous payloads further necessitates context-aware adaptability; Aco2 utilizes contextual contrastive meta-RL to autonomously extract implicit representations of unmodeled dynamic mutations triggered by different object masses and geometries \cite{jin2026autonomous}. For minimalist, lightweight hardware designs—such as a 2-DoF Differential-Drive Spatial Aerial Manipulator (DSAM)—RL policies can be synthesized to provide direct global end-effector 6-DoF pose control, overcoming the extreme underactuation that classical controllers struggle to solve \cite{deshmukh2025global}.

\subsubsection{Task-Specific Execution and High-Speed Grasping}
Different grasping tasks demand specialized algorithmic coordination to achieve centimeter-level precision. In agricultural applications like fruit picking, interaction data exhibits highly non-independent and identically distributed (non-IID) characteristics, where the dynamics of rapid approach vastly differ from those of precise contact adjustments. Sub-task imitation learning explicitly addresses this non-IID distribution, preventing policy collapse across heterogeneous phases \cite{baraban2021perception}. To mitigate cross-platform generalization issues caused by customized hardware couplings across various tasks, frameworks like Flying Hand advocate for an end-effector-centric teleoperation and policy learning paradigm \cite{he2025flying}. Finally, tasks requiring extreme velocity, such as the Swooper system, demonstrate that capturing a moving object at 1.5 m/s necessitates tightly coupled synchronization between flight control and gripper actuation, relying on progressive RL curriculum training to avoid early-stage collision penalties \cite{huang2025swooper}.

%% file: sections/uav-embodied-cooperation.tex
\section{UAV Embodied Collaboration}

UAV embodied collaboration extends intelligence from individual platforms to coordinated perception, reasoning, and physical action among multiple embodied agents. Compared with single-agent systems, collaboration enables UAVs to exploit complementary viewpoints, sensing capabilities, mobility patterns, and physical skills, allowing complex missions beyond the capability of a single platform. Recent progress has shifted collaborative UAV systems from homogeneous information sharing toward heterogeneous embodied teams, where different agents contribute specialized capabilities through coordinated perception, decision-making, and execution. This section reviews collaborative tasks, analyzes common collaboration mechanisms, and summarizes representative benchmarks and simulation platforms.

\subsection{Collaborative Tasks}

\subsubsection{Collaborative Perception}

Collaborative perception aims to construct a more complete environmental understanding by integrating observations from multiple agents. For UAV systems, aerial platforms provide wide-area coverage, flexible viewpoints, and reduced occlusion, while heterogeneous agents contribute complementary sensing capabilities. Existing studies mainly focus on feature sharing, representation alignment, and uncertainty-aware perception to overcome viewpoint, scale, and communication differences among agents.

CoCa3D investigates communication-efficient collaborative perception through intermediate feature exchange among multiple agents \cite{hu2023collaboration}. OpenCOOD-Air extends this paradigm to heterogeneous air-ground systems by addressing differences between aerial and ground observations \cite{wu2026opencood}. VL-UniTrack explores cross-view tracking through shared visual-language representations while preserving geometric information across viewpoints \cite{xu2026vl}. FusionTrack further studies multi-object tracking with arbitrary multi-UAV camera configurations through a joint tracking-association framework \cite{li2026fusiontrack}. Beyond passive information fusion, Active Metric-Semantic Mapping incorporates uncertainty-aware active sensing to determine informative observations during collaborative mapping \cite{liu2023active}. These studies demonstrate a transition from simple observation aggregation toward embodied collaborative perception, where agents actively acquire, share, and utilize complementary information.

\subsubsection{Collaborative Navigation}

Collaborative navigation coordinates observations, spatial roles, and motion decisions among multiple agents to accomplish shared navigation objectives. UAVs naturally benefit from such cooperation because different altitudes, viewpoints, and mobility patterns provide complementary spatial information. Current research extends from multi-agent navigation toward cooperative search, exploration, and heterogeneous air-ground missions.

AeroDuo investigates dual-altitude UAV cooperation, where agents with different viewpoints perform complementary navigation roles \cite{wu2025aeroduo}. CoNav-UAV further introduces structured coordination between UAVs through leader-follower navigation strategies \cite{song2026conav}. AGC-VLN establishes a training-free air-ground vision-and-language navigation baseline: the UAV renders its global bird's-eye view with the UGV pose and target position as shared markers, while the UGV performs road-level planning and the UAV runs three-dimensional spatial search \cite{zhang2026air}. LAGCN extends collaborative navigation to UAV--unmanned ground vehicle (UGV) systems by using aerial perception to guide ground movement in complex environments \cite{wang2026lagcn}. For exploration and search tasks, GoalSwarm introduces semantic coordination among multiple UAVs \cite{james2026goalswarm}, while uncertainty-aware search considers target uncertainty and information gain during decentralized exploration \cite{tabib2024decentralized}. AGOS-Agent further formulates air-ground object search as a search--handoff--verify loop, using aerial discovery, candidate projection, road-constrained ground approach, and close-range verification \cite{yu2026towards}. RACER, C$^2$-Explorer, and MOCHA study decentralized exploration under constraints from task allocation, connectivity, and communication \cite{zhou2023racer,yan2026c,cladera2024enabling}. These efforts indicate a shift from coordinated motion toward persistent information gathering, cross-view grounding, and adaptive mission execution.

\subsubsection{Collaborative Task Planning}

Collaborative task planning addresses how heterogeneous agents jointly decompose missions, allocate responsibilities, and execute complex objectives. Unlike conventional multi-agent scheduling, embodied collaboration requires explicit reasoning about agent capabilities, available resources, and physical execution constraints.

COHERENT introduces capability descriptions and execution feedback into heterogeneous robot collaboration, enabling task-level coordination between different embodiments \cite{liu2025coherent}. EMOS further explores capability-aware negotiation by allowing agents to represent their own functional properties during collaboration \cite{chen2025emos}. D-VLC considers decentralized collaboration with online capability assessment and assistance requests when local execution is insufficient \cite{zhou2026d}. MultiUAV-Plat investigates FM-based planning for UAV teams and introduces execution validation mechanisms to improve task reliability \cite{zhang2026multiuav}. These studies show a progression from predefined role assignment toward adaptive collaboration based on embodied capability understanding.

\subsubsection{Collaborative Motion Coordination}

Collaborative motion coordination focuses on generating physically consistent behaviors among multiple agents under spatial, dynamic, and communication constraints. Compared with task-level coordination, these methods directly address how agents move together and satisfy shared physical objectives.

Swarm-GPT explores FM-assisted swarm behavior generation by translating high-level descriptions into coordinated motion patterns \cite{jiao2023swarm}. FlockGPT extends language-guided coordination to flocking behaviors through geometric representations and motion constraints \cite{lykov2024flockgpt}. SwarmGPT further combines generated behaviors with optimization-based refinement for executable swarm trajectories \cite{schuck2025swarmgpt}. Beyond pure motion coordination, MRLMN investigates communication-aware UAV network coordination through multi-agent reinforcement learning (MARL) \cite{xu2026scalable}, while DGPPO and CrazyMARL study physically coupled UAV systems involving payload transportation and cable constraints \cite{choi2026safe,lorentz2025crazymarl}. These studies demonstrate that collaborative motion is gradually evolving from geometric coordination toward physically coupled multi-agent interaction.

\subsubsection{Collaborative Aerial Manipulation and Operation}

Collaborative aerial manipulation extends multi-agent cooperation from perception and motion coordination toward direct physical interaction. Such tasks require multiple UAVs or heterogeneous platforms to jointly perform manipulation, transportation, construction, or tool-use operations while handling contact uncertainty and dynamic coupling.

Aerial-AM demonstrates cooperative aerial operation in construction scenarios by coordinating UAVs with different functional roles for deposition and inspection \cite{zhang2022aerial}. FlyingToolbox further explores physical cooperation through aerial docking and tool exchange between platforms \cite{cao2025proximal}. Compared with information-level collaboration, aerial manipulation introduces stronger coupling among agents, payloads, and environments, representing an important direction toward fully embodied multi-agent systems.


\begin{table*}[!t]
\centering
\caption{Comparison of Representative Approaches to UAV Embodied Collaboration}
\label{tab:uav_embodied_collaboration}

\scriptsize
\setlength{\tabcolsep}{3.2pt}
\renewcommand{\arraystretch}{1.0}

\resizebox{\textwidth}{!}{%
\begin{tabular}{
@{}
>{\raggedright\arraybackslash}p{2.60cm}
>{\raggedright\arraybackslash}p{1.80cm}
>{\raggedright\arraybackslash}p{2.30cm}
>{\raggedright\arraybackslash}p{2.60cm}
>{\raggedright\arraybackslash}p{1.90cm}
>{\raggedright\arraybackslash}p{2.90cm}
>{\raggedright\arraybackslash}p{1.55cm}
@{}
}
\toprule
\textbf{Method} &
\textbf{Embodied Team} &
\textbf{Coordination} &
\textbf{Shared Representation} &
\textbf{FM Role} &
\textbf{Execution / Assurance} &
\textbf{Validation} \\
\midrule

\multicolumn{7}{@{}l}{\textit{Collaborative Perception}} \\
CoCa3D~\cite{hu2023collaboration}
& Camera agents
& Intermediate fusion
& Sparse features
& N/A
& Comm.-aware selection
& Real/sim. \\

OpenCOOD-Air~\cite{wu2026opencood}
& UAV--UGV
& Hetero. feature fusion
& Cross-domain features
& N/A
& Spatial conversion + correction
& Simulation \\

VL-UniTrack~\cite{xu2026vl}
& UAV--camera
& Cross-view encoding
& VL features + geometric prompts
& VLM prompting
& Confidence-based distillation
& Benchmark \\

FusionTrack~\cite{li2026fusiontrack}
& Multi-UAV
& Joint tracking + association
& Tracklet/\allowbreak{}identity pools
& N/A
& View-aware clustering + filtering
& Real benchmark \\

Active Metric-Semantic Mapping~\cite{liu2023active}
& Aerial robots
& Active mapping
& Uncertain metric-semantic maps
& N/A
& Uncertainty-driven selection
& Real \\

\midrule
\multicolumn{7}{@{}l}{\textit{Collaborative Navigation}} \\
AeroDuo~\cite{wu2025aeroduo}
& Multi-UAV
& Altitude-specific roles
& Cross-view geometry
& VLM reasoner
& Low-altitude avoidance
& Simulation \\

AGC-VLN~\cite{zhang2026air}
& UAV--UGV
& Shared BEV map
& Viewpoint-invariant annotations
& VLM reasoner
& Projection + closed-loop control
& Real/sim. case \\

CoNav-UAV~\cite{song2026conav}
& Multi-UAV
& Stackelberg leader-follower
& Lightweight coordinate queue
& VLM reasoner
& Collision-free control
& Simulation \\

LAGCN~\cite{wang2026lagcn}
& UAV--UGV
& Aerial diffusion guidance
& Semantic BEV map
& N/A
& Risk-guided generation
& Real/sim. \\

GoalSwarm~\cite{james2026goalswarm}
& Multi-UAV
& Decentralized semantic coordination
& Shared semantic map
& FM perception
& Decentralized selection
& Simulation \\

Uncertainty-Aware Search~\cite{tabib2024decentralized}
& Multi-UAV
& Decentralized active search
& Shared target beliefs
& N/A
& Uncertainty-driven planning
& Real/sim. \\

AGOS-Agent~\cite{yu2026towards}
& UAV--UGV
& Search--handoff--verify protocol
& Candidate regions + task state
& Role-conditioned VLM
& Projection, road planning + validation
& Simulation benchmark \\

RACER~\cite{zhou2023racer}
& Multi-UAV
& Decentralized exploration and allocation
& Spatial task decomposition
& N/A
& Coverage planning
& Simulation \\

C$^2$-Explorer~\cite{yan2026c}
& Multi-UAV
& Connectivity-aware allocation
& Connectivity graph task units
& N/A
& Constrained exploration
& Real/sim. \\

MOCHA~\cite{cladera2024enabling}
& UAV--UGV
& Gossip communication
& Gossip-propagated information
& N/A
& Comm.-aware planning
& Real/sim. \\

\midrule
\multicolumn{7}{@{}l}{\textit{Collaborative Task Planning}} \\
COHERENT~\cite{liu2025coherent}
& UAV--quadruped--arm
& Centralized PEFA loop
& Capabilities + execution feedback
& Task planner
& Feasibility checks + replanning
& Simulation \\

EMOS~\cite{chen2025emos}
& Heterogeneous robots
& Multi-agent negotiation
& Robot resumes + task state
& Task planner
& Capability-aware consensus
& Simulation \\

D-VLC~\cite{zhou2026d}
& Heterogeneous robots
& Async. decentralized execution
& Local maps + updates
& VLM reasoner
& Capability checks + assistance
& Simulation \\

MultiUAV-Plat~\cite{zhang2026multiuav}
& Multi-UAV
& LLM tool-interaction planning
& Role-based task state
& Task planner
& Hidden validation checks
& Simulation \\

\midrule
\multicolumn{7}{@{}l}{\textit{Collaborative Motion Coordination}} \\
Swarm-GPT~\cite{jiao2023swarm}
& Multi-UAV
& LLM with safe motion planner
& Time-indexed waypoints
& Creative generator
& Optimization-based correction
& Real/sim. \\

FlockGPT~\cite{lykov2024flockgpt}
& Multi-UAV
& Language-guided flocking
& SDF geometry
& Creative generator
& Flocking + repulsion
& Real/sim. \\

SwarmGPT~\cite{schuck2025swarmgpt}
& Multi-UAV
& LLM + distributed optimization
& Waypoints/\allowbreak{}dense trajectories
& Creative generator
& Optimization safety filter
& Real/sim. \\

MRLMN~\cite{xu2026scalable}
& UAV network
& Grouped MARL deployment
& Network/\allowbreak{}policy state
& Distillation teacher
& Key-agent constraints
& Simulation \\

DGPPO~\cite{choi2026safe}
& Payload UAVs
& Shared-policy MARL
& Local graph + waypoints
& N/A
& DGCBF policy + tracking
& Real/sim. \\

CrazyMARL~\cite{lorentz2025crazymarl}
& Payload UAVs
& Decentralized motor-level MARL
& Local observations and cable states
& N/A
& Motor control under slack/\allowbreak{}taut states
& Real/sim. \\

\midrule
\multicolumn{7}{@{}l}{\textit{Collaborative Aerial Manipulation and Operation}} \\
Aerial-AM~\cite{zhang2022aerial}
& Deposition/\allowbreak{}inspection UAVs
& Construction/\allowbreak{}inspection roles
& Print-geometry feedback
& N/A
& MPC deposition + in-flight checks
& Real \\

FlyingToolbox~\cite{cao2025proximal}
& Toolbox/\allowbreak{}manipulator MAVs
& Vertical-stack proximal operation
& Docking and tool state
& N/A
& Precise midair docking and tool exchange
& Real \\

\bottomrule
\end{tabular}%
}

\parbox{\textwidth}{
\vspace{2pt}
\scriptsize
\textit{Notes:} SDF: signed distance function; DGCBF: decentralized graph
control barrier function; MAV: micro aerial vehicle; N/A: no primary FM component.
}
\end{table*}

\subsection{Methodological Analysis}

The collaborative tasks discussed above share several fundamental challenges: how agents reason jointly, how heterogeneous information is represented and exchanged, and how collaborative decisions are grounded into physically executable behaviors. These aspects determine whether multi-agent systems can evolve from information sharing toward reliable embodied cooperation.

\subsubsection{The Role of Foundation Models in Collaboration}

FM introduces new opportunities for collaborative cognition by providing semantic understanding, reasoning, and transferable knowledge. Existing studies mainly explore several roles, including task planning, navigation reasoning, behavior generation, and knowledge transfer.

At the task level, COHERENT introduces FM-based planning with execution feedback to coordinate heterogeneous robotic teams \cite{liu2025coherent}. EMOS further explores capability-aware collaboration by allowing embodied agents to represent their own capabilities during task negotiation \cite{chen2025emos}. For navigation-oriented collaboration, AeroDuo uses multimodal reasoning to associate language instructions with observations from different UAV viewpoints \cite{wu2025aeroduo}, while AGC-VLN uses a frozen VLM as a navigation reasoner for shared-map construction and road-path planning \cite{zhang2026air}. AGOS-Agent extends role-conditioned VLM use to a structured search--handoff--verify protocol, separating semantic decisions from geometric projection, road planning, and action validation \cite{yu2026towards}. Beyond direct online reasoning, Swarm-GPT and FlockGPT investigate FM-based generation of coordinated swarm behaviors, where generated outputs are refined by downstream motion modules \cite{jiao2023swarm,lykov2024flockgpt}. MRLMN adopts a different strategy by using FM as a teacher to transfer high-level coordination knowledge into MARL policies \cite{xu2026scalable}.

These studies reveal a common design principle: FM mainly provides high-level cognitive capabilities, while specialized planners, controllers, or learned policies remain responsible for reliable physical execution.

\subsubsection{Information Sharing and Alignment Strategies}

Information sharing determines how multiple agents integrate partial observations and internal states under heterogeneous sensing conditions. Existing approaches mainly follow feature-space alignment and unified representation design.

Feature-space alignment directly exchanges perceptual information among agents. CoCa3D investigates communication-efficient collaborative perception through selective intermediate feature sharing \cite{hu2023collaboration}. OpenCOOD-Air extends this paradigm to heterogeneous air-ground systems by considering differences in viewpoint, scale, and spatial structure \cite{wu2026opencood}. Such approaches preserve detailed perceptual information but require effective alignment mechanisms when sensing modalities and embodiments become increasingly diverse.

Unified representation approaches instead transform heterogeneous observations into common task-oriented spaces. VL-UniTrack introduces shared visual-language representations for cross-view interaction while maintaining geometric information from different viewpoints \cite{xu2026vl}. LAGCN uses bird's-eye-view (BEV) semantic representations to connect aerial perception with ground navigation \cite{wang2026lagcn}. AGC-VLN uses an annotated BEV image as a viewpoint-invariant interface between aerial perception and ground execution, whereas AGOS-Agent exchanges structured candidate regions and coordination state for search handoff \cite{zhang2026air,yu2026towards}. FusionTrack couples temporal tracklet features with cross-view identity representations for multi-UAV association \cite{li2026fusiontrack}. D-VLC further explores structured representation sharing for decentralized heterogeneous collaboration \cite{zhou2026d}. These methods improve scalability and interoperability, but the abstraction process may remove task-specific details.

Therefore, the key challenge is not simply increasing communication volume, but determining what information should be shared, at what abstraction level, and under which task constraints.

\subsubsection{Safety Assurance Mechanisms}

Collaborative behaviors must satisfy geometric, dynamic, communication, and physical constraints before execution. Existing approaches introduce safety through explicit representations, optimization-based refinement, and learning-based mechanisms.

FlockGPT uses geometric representations to describe desired swarm structures and combines them with motion constraints for collision avoidance \cite{lykov2024flockgpt}. Swarm-GPT generates coordinated motion patterns and employs safety-aware refinement before execution \cite{jiao2023swarm}. SwarmGPT further integrates optimization-based filtering to improve trajectory feasibility for generated behaviors \cite{schuck2025swarmgpt}.

For physically coupled tasks, DGPPO incorporates safety constraints into learned policies for cooperative payload transportation \cite{choi2026safe}. Other approaches consider task-specific constraints beyond physical collision avoidance. LAGCN incorporates semantic risk into navigation generation \cite{wang2026lagcn}, while MRLMN considers communication constraints during multi-agent coordination \cite{xu2026scalable}.

AGC-VLN and AGOS-Agent place deterministic geometric projection, road-feasible planning, and action validation between foundation-model decisions and physical execution \cite{zhang2026air,yu2026towards}. This limits the propagation of semantic errors, but it does not remove the risk of incorrect target localization or false ground verification. AGOS-Agent explicitly exposes this failure mode through its distinction between aerial candidate guidance and ground verification accuracy.

These studies demonstrate that safety should be integrated into the collaboration architecture rather than treated as an independent post-processing module. Future systems will require stronger connections among semantic reasoning, predictive models, and safety-critical execution.

\subsection{Benchmark and Simulation Platforms}

Evaluation of collaborative UAV systems spans perception, reasoning, mission execution, and physical coordination. Existing resources can be divided into collaborative perception benchmarks, task-level mission benchmarks, and simulation platforms.

\subsubsection{Collaborative Perception Benchmarks and Datasets}

Collaborative perception benchmarks mainly evaluate whether multiple agents can effectively exchange and integrate observations. AirCopBench provides a comprehensive benchmark for multi-UAV collaborative perception and reasoning under diverse sensing conditions \cite{zha2026aircopbench}. V2U4Real introduces real-world air-ground collaborative perception data with dynamic environments and non-ideal sensing conditions \cite{li2026v2u4real}. AirV2X complements real-world datasets through simulated air-ground interaction scenarios and supports evaluation of collaborative perception tasks \cite{gao2025airv2x}. VVSim adds a large-scale aerial-ground dataset with about 61k annotated frames, 255k LiDAR sweeps, 3.5M multimodal images, 19 traffic interaction scenarios, five weather conditions, and 11 scenes; its VVFormer model provides a unified benchmark for heterogeneous cross-view fusion \cite{zhu2026vvsim}. MDMOT targets a different perception problem, namely multi-object tracking under arbitrary and dynamically changing multi-UAV camera configurations. It contains 122k frames and 2.48M annotations from a coordinated drone fleet, and evaluates cross-view association in both overlapping and non-overlapping views \cite{li2026fusiontrack}.

\subsubsection{Collaborative Task and Mission Benchmarks}

Task-level benchmarks evaluate whether embodied agents can understand capabilities, coordinate responsibilities, and complete complex missions. The COHERENT Benchmark studies heterogeneous robot collaboration involving different embodiments and task requirements \cite{liu2025coherent}. Habitat-MAS extends evaluation toward multi-agent embodied reasoning across navigation, perception, and manipulation scenarios \cite{chen2025emos}. MultiUAV-Plat focuses on UAV-only collaboration and introduces validation mechanisms for evaluating task completion reliability \cite{zhang2026multiuav}. AGOS-Bench specializes in urban air-ground object search: its AGOS-Dataset contains 7,700 training and validation episodes across five CARLA towns, plus 210 held-out test episodes in an unseen town, with easy, medium, and hard settings. It evaluates not only success, oracle success, path efficiency, and navigation error, but also aerial search efficiency, candidate guidance, ground verification accuracy, and decision-step cost \cite{yu2026towards}. The benchmark therefore separates wide-area discovery, cross-view handoff, and close-range verification instead of collapsing them into a single success score.

\subsubsection{Collaborative Simulation Platforms}

Simulation platforms provide controllable environments for closed-loop evaluation of collaborative behaviors. AirSimAG supports heterogeneous air-ground missions including mapping, planning, tracking, and exploration \cite{cui2026airsimag}. CARLA-Air integrates UAV and ground vehicle simulation within urban environments to support air-ground collaboration studies \cite{zeng2026carla}. HERCULES further provides a unified environment for heterogeneous perception, mapping, and exploration \cite{garimella2026hercules}. AGC-VLN uses CARLA-Air to evaluate a training-free air-ground navigation baseline, while AGOS-Bench uses the same class of urban simulation infrastructure to evaluate search, handoff, and verification as a closed-loop task \cite{zhang2026air,yu2026towards}.

Together, these resources indicate a shift from evaluating isolated collaborative modules toward complete collaborative perception-reasoning-action loops.

%% file: sections/conclusion.tex
\section{Challenges and Future Directions}
\label{sec:challenges}

Despite rapid progress in perception, planning, interaction, and collaboration, current UAV embodied intelligence remains largely optimized for bounded missions and predefined operating conditions. Achieving more general aerial autonomy requires addressing several capability gaps related to long-duration operation, physical understanding, adaptation, and continual capability development.

\subsection{Challenges}

\subsubsection{Long-Horizon Aerial Autonomy}

UAVs often operate over large spatial ranges and extended durations, where observations are acquired across changing viewpoints, scales, and environmental conditions. Maintaining coherent understanding requires more than increasing model context length; an EI UAV must selectively retain important information, associate observations across time and space, and relate current decisions to previous mission states. Developing effective memory and spatiotemporal reasoning mechanisms is therefore essential for persistent aerial autonomy.

\subsubsection{Flight-Grounded Physical Reasoning}

An EI UAV must understand not only the external environment but also the coupling among its embodiment, dynamics, payloads, and physical interactions. Actions influence future observations and physical states, while disturbances and object properties constrain feasible behaviors. Therefore, UAV intelligence requires predictive physical reasoning that can anticipate environmental evolution, interaction outcomes, and the consequences of alternative actions, moving beyond semantic understanding toward Physical AI for aerial systems.

\subsubsection{Changing Flight Conditions Adaptation}

Real-world UAV operation is inherently non-stationary, with variations in weather, illumination, sensing quality, payload, battery state, and vehicle configuration. An EI UAV should adapt its perception, planning, and control behaviors according to changing flight conditions without relying on extensive retraining. The key challenge is to achieve rapid online adaptation from limited experience while maintaining stability, safety, and physical feasibility.

\subsubsection{Experience-Driven Capability Evolution}

Beyond adapting existing behaviors, future EI UAVs should progressively improve their capabilities through accumulated experience. Long-term operation generates valuable knowledge about environments, tasks, and platform-specific behaviors, which should be transformed into reusable skills, workflows, and decision strategies. The fundamental challenge is to autonomously consolidate, refine, and reorganize capabilities while preserving previously acquired knowledge.

\begin{figure}[htbp]
    \centering
    \includegraphics[width=\columnwidth]{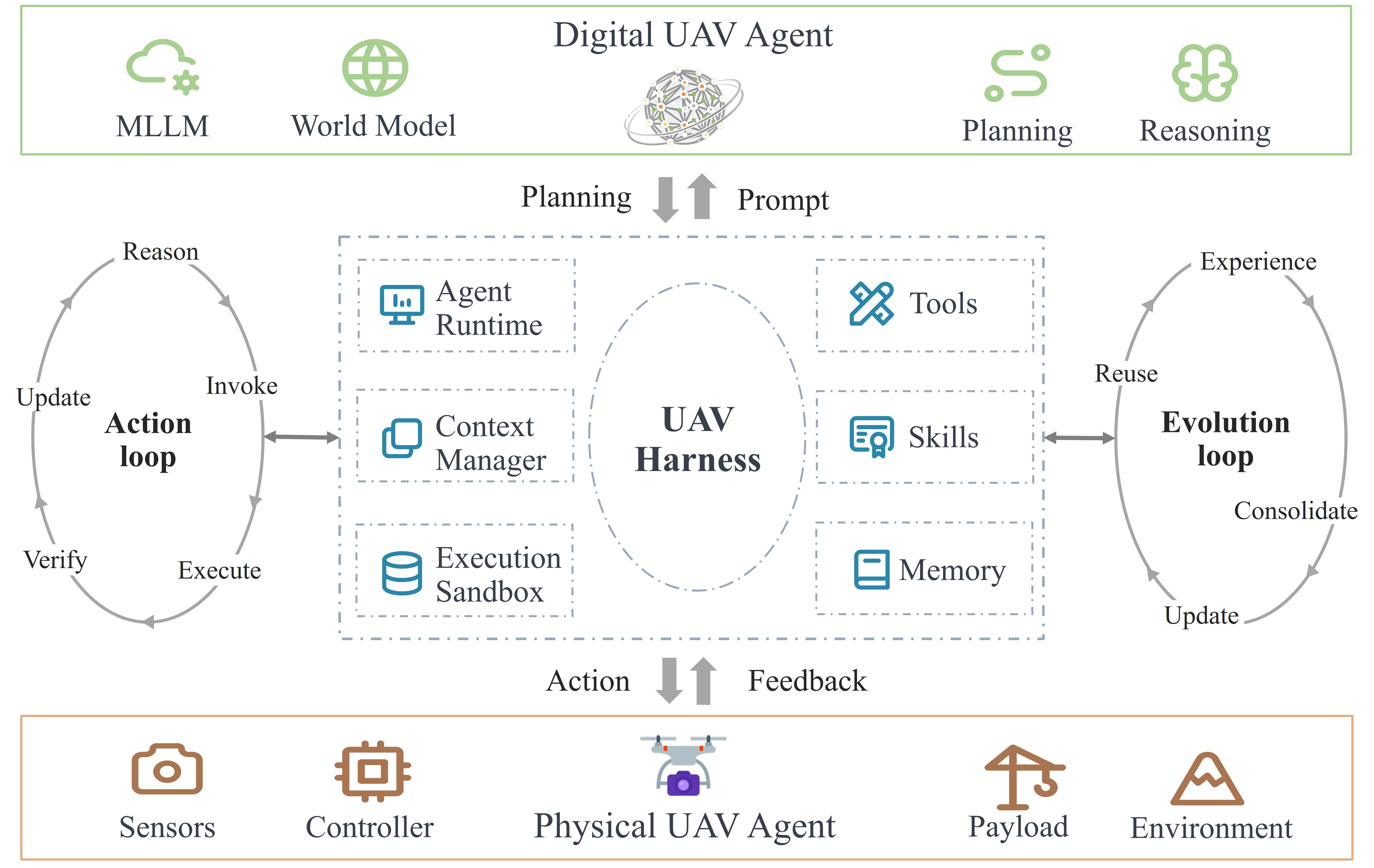}
    \caption{Harnessing physical-digital AI agents for UAV embodied intelligence. The Harness Layer coordinates digital reasoning with physical execution through runtime orchestration, while accumulated interaction experience is consolidated into reusable memory, knowledge, skills, and workflows to support persistent capability development.}
    \label{fig:harness_framework}
\end{figure}

\subsection{Harnessing Physical-Digital AI Agents}

The challenges above are not isolated problems that can be addressed by improving individual perception, planning, or control modules alone. Long-horizon autonomy, physical reasoning, online adaptation, and capability evolution all require digital intelligence to remain persistently coupled with physical execution and feedback. Recent advances in robot harnesses suggest a shift from model-centric intelligence toward system-level embodied agents, where runtime infrastructures coordinate models, tools, skills, memory, and physical resources to support persistent interaction and capability development \cite{lee2026meta,chen2026show}. Inspired by this emerging direction, we envision a UAV-specific Harness for aerial embodied intelligence. As illustrated in Fig.~\ref{fig:harness_framework}, PD-AI agents realize this coupling through a Digital UAV Agent, a UAV Harness, and a Physical UAV Agent. The digital agent provides multimodal understanding, world modeling, reasoning, and planning, while the physical agent grounds these decisions through sensing, control, payload operation, and environmental interaction. The Harness bridges these two levels by translating high-level decisions into executable actions and returning physical feedback to subsequent reasoning.

The Harness centers on two coupled loops. In the \textbf{action loop}, agent runtime and context management organize task execution, invoke appropriate tools and skills, and pass candidate actions through an execution sandbox before physical execution; observed outcomes are then verified and used to update the current task state. In the \textbf{evolution loop}, successful and failed interaction experience is consolidated into memory and reusable skills, which can be recalled and refined in later missions. Together, these loops enable the UAV to move beyond one-shot reasoning toward persistent closed-loop execution and experience-driven capability evolution.

\section{Conclusion}
\label{sec:conclusion}

UAV embodied intelligence is evolving from task-specific perception and control toward integrated systems that couple semantic understanding, memory, physical reasoning, planning, and closed-loop execution. A unified view of this field requires connecting capability development with system realization, including human intent understanding, embodied self-awareness, environment understanding, task planning, and execution across the Embodiment, Brain, Cerebellum, Interaction, and Harness layers. Recent progress in morphology, embodied perception, world models, planning, vision-language navigation, aerial manipulation, collaboration, and benchmarks further shows a clear shift toward tighter coupling between high-level intelligence and physical interaction. Harnessing parallel physical-AI agents provides a system-level pathway for this transition by linking reasoning and capability organization with sensing, dynamics, control, and real-world feedback, supporting more general, adaptive, and continuously evolving aerial autonomy.